\documentclass[10pt,journal,compsoc]{IEEEtran}
\usepackage{graphicx}
\usepackage{amsmath}
\usepackage{pgfplots}
\usepackage{multirow}
\usepackage[switch]{lineno} 

\ifCLASSOPTIONcompsoc
  \usepackage[nocompress]{cite}
\else
  \usepackage{cite}
\fi
\ifCLASSINFOpdf
\else
\fi

\begin{document}
%
\title{Incorporating Network Built-in Priors in Weakly-supervised Semantic Segmentation }
%
%
%
%

\author{Fatemeh~Sadat~Saleh, 
        Mohammad~Sadegh~Aliakbarian, 
        Mathieu~Salzmann, 
        Lars~Petersson, 
        Jose~M.~Alvarez, 
        and~Stephen~Gould
\IEEEcompsocitemizethanks{\IEEEcompsocthanksitem The authors are with the College of Engineering and Computer Science,
Australian National University and Data61-CSIRO, Australia.
\protect\\
Contact E-mail: fatemehsadat.saleh@data61.csiro.au
\IEEEcompsocthanksitem Mathieu Salzmann is with CVLab, EPFL, Switzerland}
}

\IEEEtitleabstractindextext{%
\begin{abstract}
Pixel-level annotations are expensive and time consuming to obtain. Hence,  weak supervision using only image tags could have a significant impact in semantic segmentation. Recently, CNN-based methods have proposed to fine-tune pre-trained networks using image tags. Without additional information, this leads to poor localization accuracy. This problem, however, was alleviated by making use of objectness priors to generate foreground/background masks. Unfortunately these priors either require pixel-level annotations/bounding boxes, or still yield inaccurate object boundaries. Here, we propose a novel method to extract accurate masks from networks pre-trained for the task of object recognition, thus forgoing external objectness modules. We first show how foreground/background masks can be obtained from the activations of higher-level convolutional layers of a network. We then show how to obtain multi-class masks by the fusion of foreground/background ones with information extracted from a weakly-supervised localization network. Our experiments evidence that exploiting these masks in conjunction with a weakly-supervised training loss yields state-of-the-art tag-based weakly-supervised semantic segmentation results.
\end{abstract}

}

\maketitle

\IEEEdisplaynontitleabstractindextext

%
\IEEEpeerreviewmaketitle

\ifCLASSOPTIONcompsoc
\IEEEraisesectionheading{\section{Introduction}\label{sec:introduction}}
\else
\section{Introduction}
\label{sec:introduction}
\fi
\IEEEPARstart{S}{emantic} scene segmentation, i.e., assigning a class label to every pixel in an input image, has received growing attention in the computer vision community, with accuracy greatly increasing over the years~\cite{FCN,deconvnet,DeepLab,crf-rnn,zoomout,deephierarchy}. In particular, fully-supervised approaches based on Convolutional Neural Networks (CNNs) have recently achieved impressive results~\cite{farabetSeg,FCN,DeepLab,deconvnet,crf-rnn}. Unfortunately, these methods require large amounts of training images with pixel-level annotations, which are expensive and time-consuming to obtain.
Weakly-supervised techniques have therefore emerged as a solution to address this limitation~\cite{graphbasedseg,tellme,multiimagemodel,segmentvarious,whatsthepoint,weaklyandsemi,MIL,cliqueseg,learnTosegwithimagelevelanno,STC,SEC}. These techniques rely on a weaker form of training annotations, such as, from weaker to stronger levels of supervision, image tags~\cite{MIL,whatsthepoint,fromimgtopixel,CCNN,learnTosegwithimagelevelanno,STC,SEC}, information about object sizes~\cite{CCNN}, labeled points or squiggles~\cite{whatsthepoint} and labeled bounding boxes~\cite{weaklyandsemi,boxsup,boxeccv}. In the current Deep Learning era, existing weakly-supervised methods typically start from a network pre-trained on an object recognition dataset (e.g., ImageNet~\cite{ILSVRC}) and fine-tune it using segmentation losses defined according to the weak annotations at hand~\cite{fromimgtopixel,CCNN,weaklyandsemi,whatsthepoint,MIL}.

In this paper, we are particularly interested in exploiting one of the weakest levels of supervision, i.e., image tags, which are rather inexpensive attributes to annotate and thus more common in practice (e.g., Flickr~\cite{flickr}). Image tags simply determine which classes are present in the image without specifying any other information, such as the location of the objects. In this extreme setting, a naive weakly-supervised segmentation algorithm will typically yield poor localization accuracy. Therefore, recent works~\cite{fromimgtopixel,whatsthepoint,learnTosegwithimagelevelanno} have proposed to make use of objectness priors~\cite{objectness,BING,MCG,CPMC}, which provide each pixel with a probability of being an object. In particular, these methods have exploited existing objectness algorithms, such as~\cite{objectness,BING,MCG}, with the drawback of introducing external sources of potential error. Furthermore,~\cite{objectness} typically only yields a rough foreground/background estimate, and~\cite{BING,MCG} rely on additional training data with pixel-level annotations. 

Here, by contrast, we introduce a Deep Learning approach to weakly-supervised semantic segmentation where the localization information is directly extracted from networks pre-trained for the task of object recognition. Our approach relies on the following intuition: One can expect that a network trained to recognize objects in images extracts features that focus on the objects themselves, and thus has hidden layers with units firing up on foreground objects, but not on background regions. A similar intuition was also recently explored for object detection~\cite{ObjectEmerge} and localization~\cite{ObjectLocalForFree}, which inspired the contemporary weakly-supervised semantic segmentation work~\cite{SEC}. In this paper, we propose to exploit this intuition to generate (i) a foreground/background mask; and (ii) a multi-class mask. 

\begin{figure*}[!t]
\centering
\includegraphics[width=1\textwidth]{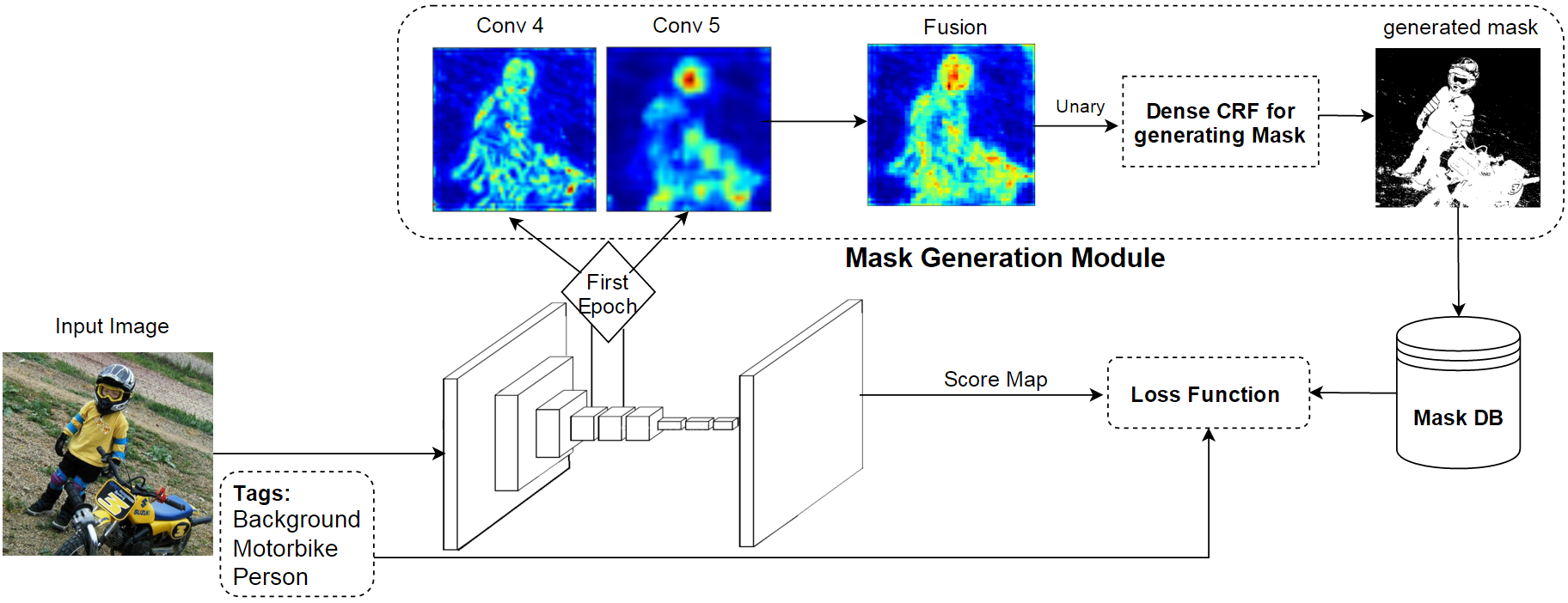}
\caption{Overview of our weakly-supervised network with built-in foreground/background prior.}
\label{fig:intro}
\end{figure*}

\begin{figure*}[!t]
\centering
\includegraphics[width=1\textwidth]{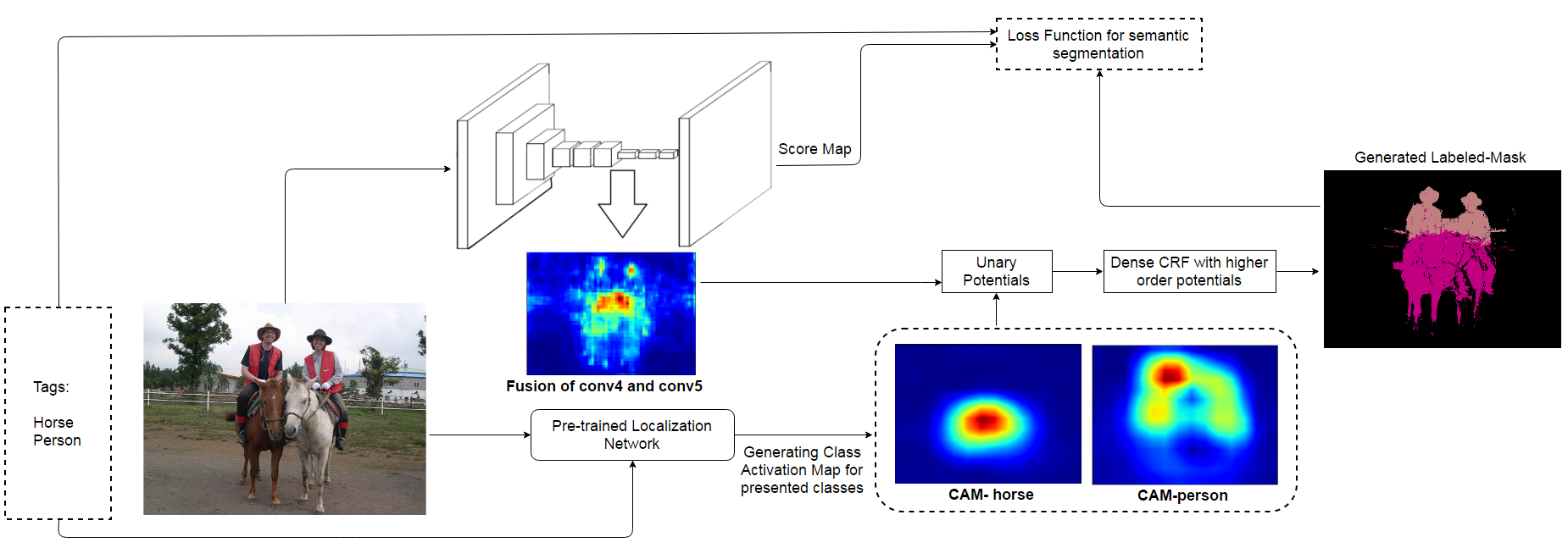}
\caption{Overview of our weakly-supervised network with multi-class masks.}
\label{fig:proposed_labeled}
\end{figure*}

More specifically, starting from a fully-convolutional network pre-trained on ImageNet, we propose to extract a foreground/background mask by directly exploiting the unit activations of some of the hidden layers in the network.
In particular, as illustrated in Fig.~\ref{fig:intro}, we focus on the fourth and fifth convolution layers of the VGG-16 pre-trained network~\cite{VGG16}, which provide higher-level information than the first three layers, such as highlighting complete objects or object parts. Note that the resulting masks can also be thought of as a form of objectness measure. While effective, this approach only reasons about foreground/background, without explicitly considering the different foreground classes. To address this, we propose to make use of a pre-trained localization network, which specifically provides information about the location of different object classes. We then show how this information can be combined with the previous fusion-based strategy, as illustrated in Fig.~\ref{fig:proposed_labeled}, to obtain class-wise pixel probabilities. In both the foreground/background and multi-class cases, the final masks are obtained by making use of a fully-connected Conditional Random Field (CRF) with higher-order terms to smooth the initial pixelwise probabilities. In particular, we propose to make use of the crisp boundary detection method of~\cite{Crisp} to generate our higher-order terms. 

We then show how these two types of masks can be incorporated in a weakly-supervised loss to train a Deep Network for the task of semantic segmentation using only image tags as ground-truth annotations. Ultimately, since our masks are directly extracted from pre-trained networks, our approach can be thought of as a weakly-supervised segmentation network with built-in foreground/background, or multi-class prior.

We demonstrate the benefits of our approach on Pascal VOC 2012~\cite{pascal}, which is the most popular dataset for weakly-supervised semantic segmentation. 
Our experiments show that our approach outperforms the state-of-the-art methods that use image tags only, and even some methods that leverage additional supervision, such as object size information~\cite{CCNN} and point supervision~\cite{whatsthepoint}. To demonstrate the generality of our approach, we also report results on two other challenging datasets: YouTube Objects~\cite{yto} and Microsoft COCO~\cite{coco}. To the best of our knowledge, this represents the first attempt at performing weakly-supervised semantic segmentation on MS COCO.

This paper is an extended version of our conference paper~\cite{OurECCV}. In particular, while our previous work focused on foreground/background masks, here, we introduce an approach to generating class-specific masks and employing them for weakly-supervised semantic segmentation. Furthermore, we introduce new higher-order terms in our CRF by exploiting the crisp boundary detection framework~\cite{Crisp}. Finally, in addition to producing state-of-the-art results, our experiments provide a thorough evaluation of the different components of our model.

\section{Related Work}
Weakly-supervised semantic segmentation has attracted a lot of attention, because it alleviates the painstaking process of manually generating pixel-level training annotations. Over the years, great progress has been made~\cite{tellme,multiimagemodel,segmentvarious,whatsthepoint,weaklyandsemi,MIL,fromimgtopixel,CCNN,boxsup,structuredlearning,learnTosegwithimagelevelanno,STC,SEC,augment_feedback,distinct}. In particular, recently, Convolutional Neural Networks (CNNs) have been applied to the task of weakly-supervised segmentation with great success. In this section, we discuss these CNN-based approaches, which are the ones most related to our work.

The work of~\cite{MIL} constitutes the first method to consider fine-tuning a CNN pre-trained for object recognition, using image-level tags only, within a weakly-supervised segmentation context.
This approach relies on a Multiple Instance Learning (MIL) loss to account for image tags during training. While this loss improves segmentation accuracy over a naive baseline, this accuracy remains relatively low, due to the fact that no other prior than image tags is employed. By contrast,~\cite{weaklyandsemi} incorporates an additional prior in the MIL framework in the form of an adaptive foreground/background bias. This bias significantly increases accuracy, which~\cite{weaklyandsemi} shows can be further improved by introducing stronger supervision, such as labeled bounding boxes. Importantly, however, this bias is data-dependent and not trivial to re-compute for a new dataset. Furthermore, the results remain inaccurate in terms of object localization. In~\cite{CCNN}, weakly-supervised segmentation is formulated as a constrained optimization problem, and an additional prior modeling the size of objects is introduced. This prior relies on thresholds determining the percentage of the image area that certain classes of objects can occupy, which again is problem-dependent. More importantly, and as in~\cite{weaklyandsemi}, the resulting method does not exploit any information about the location of objects, and thus yields poor localization accuracy.

To overcome this weakness, some approaches~\cite{fromimgtopixel,whatsthepoint,learnTosegwithimagelevelanno,augment_feedback} have proposed to exploit the notion of objectness. In particular,~\cite{fromimgtopixel} makes use of a post-processing step that smoothes initial segmentation results using the object proposals obtained by BING~\cite{BING} or MCG~\cite{MCG}. While it improves localization, being a post-processing step, this procedure is unable to recover from some mistakes made by the initial segmentation. By contrast,~\cite{whatsthepoint,learnTosegwithimagelevelanno} directly incorporate an objectness score~\cite{objectness,MCG} in their loss function. ~\cite{augment_feedback} also uses these objectness methods to generate segmentation masks and train the semantic segmentation network iteratively. While accounting for objectness when training the network indeed improves segmentation accuracy, the whole framework depends on the success of the external objectness module, which, in practice, only produces a coarse heat map and does not accurately determine the location and shape of the objects (as evidenced by our experiments). 
Note that BING and MCG have been trained from PASCAL \textit{train} images with full pixel-level annotations or bounding boxes, and thus~\cite{fromimgtopixel,learnTosegwithimagelevelanno,augment_feedback} inherently make use of stronger supervision than our approach.
Instead of objectness, the method in~\cite{STC} relies on DRIF saliency maps~\cite{DRIF}. These saliency maps are employed to train a simple network from Flickr images, whose output then serves to train two other networks using more complicated Pascal VOC images. Note that, again, the DRIF method requires bounding boxes in its training stage, thus inherently making use of additional supervision. Recently,~\cite{distinct} tried to produce class-specific saliency maps based on the derivatives of the class scores w.r.t. the input image that provides some localization cues for segmentation. The method of~\cite{motion} also uses motion cues of weakly annotated videos to segment images with a subset of the PASCAL VOC classes.
Here, instead of relying on an external objectness or saliency method, we leverage the intuition that, within its hidden layers, a network pre-trained for object recognition should already have learned to focus on the objects themselves. This lets us generate a foreground/background mask directly from the information built into the network, which we empirically show provides a more accurate object localization prior.

Beyond foreground/background masks, the method of the contemporary work~\cite{SEC} exploits the output of the same localization network~\cite{loc} as us, but directly in a new composite loss function for weakly-supervised semantic segmentation. While effective, the method suffers from the fact that localization of some classes is inaccurate. By contrast, here, we combine our built-in foreground/background mask with information from the localization network, thus obtaining more accurate multi-class masks. As evidenced by our experiments, these more robust masks yield more accurate semantic segmentation results.

\section{Our Method}
In this section, we introduce our weakly-supervised semantic segmentation framework. First, we present our approach to extracting masks, either foreground/background or multi-class, directly from a network pre-trained for object recognition. We then introduce our weakly-supervised learning algorithms that leverage these foreground/background and multi-class masks.

\subsection{Built-in Prior Models}
Given an image, our goal is to automatically extract a mask that indicates which regions correspond to either foreground/background or specific classes.
The central idea of our approach is to rely on networks that have been pre-trained for object recognition. Intuitively, we expect that such networks have learned to focus on the objects themselves, and their parts, rather than on background regions. Below, we show how we can exploit this intuition to extract  foreground/background masks, as well as multi-class ones.

\begin{figure*}[t!]
\centering
\tiny
\begin{tabular}{c c c c c c c c c c  }
Image & \(1^{st}\) Conv. & \(2^{nd}\) Conv. & \(3^{rd}\) Conv. & \(4^{th}\) Conv. & \(5^{th}\) Conv. & Fusion & Our mask & Our mask & G.T \\
& & & & & & & & +higher order & \\

\includegraphics[width=.08\textwidth]{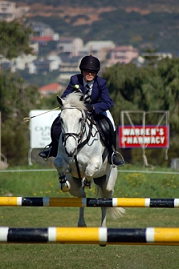} &
\includegraphics[width=.08\textwidth]{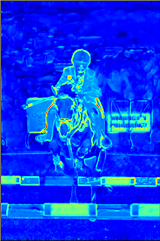} &
\includegraphics[width=.08\textwidth]{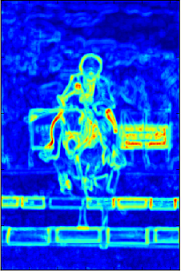} &
\includegraphics[width=.08\textwidth]{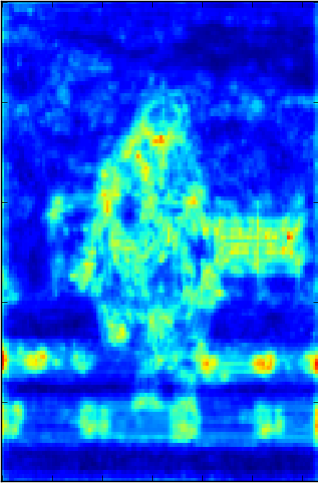} &
\includegraphics[width=.08\textwidth]{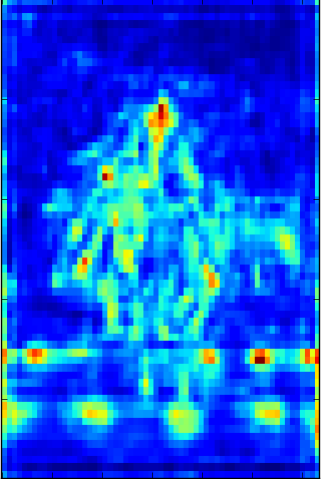} &
\includegraphics[width=.08\textwidth]{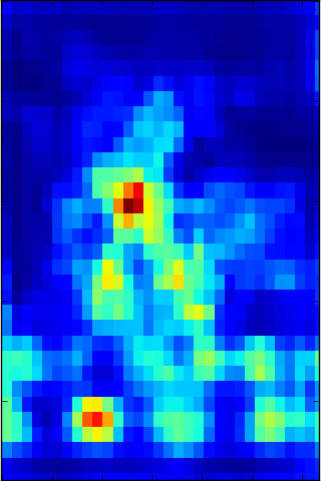} &
\includegraphics[width=.08\textwidth]{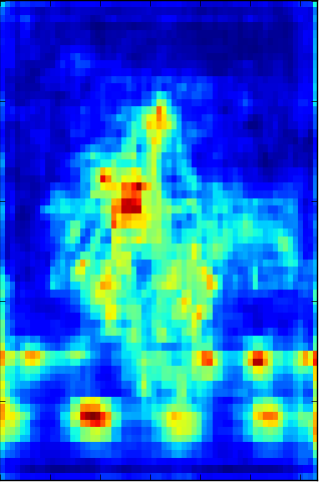} &
\includegraphics[width=.08\textwidth]{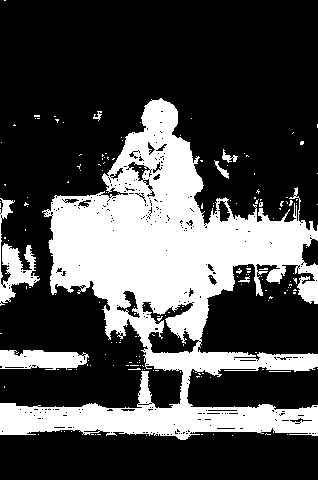} &
\includegraphics[width=.08\textwidth]{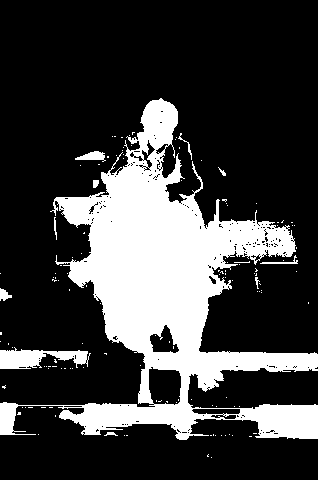} &
\includegraphics[width=.08\textwidth]{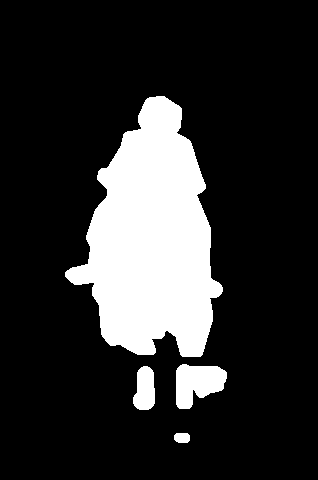} \\

\includegraphics[width=.08\textwidth]{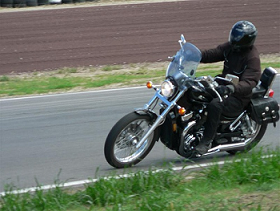} &
\includegraphics[width=.08\textwidth]{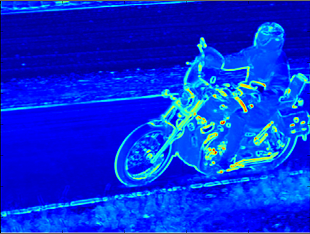} &
\includegraphics[width=.08\textwidth]{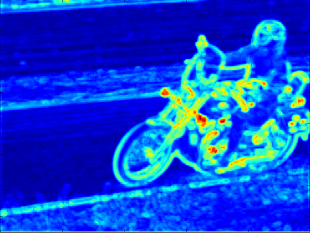} &
\includegraphics[width=.08\textwidth]{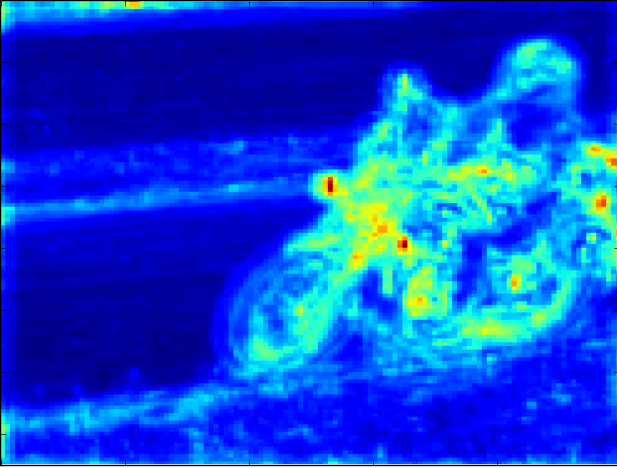} &
\includegraphics[width=.08\textwidth]{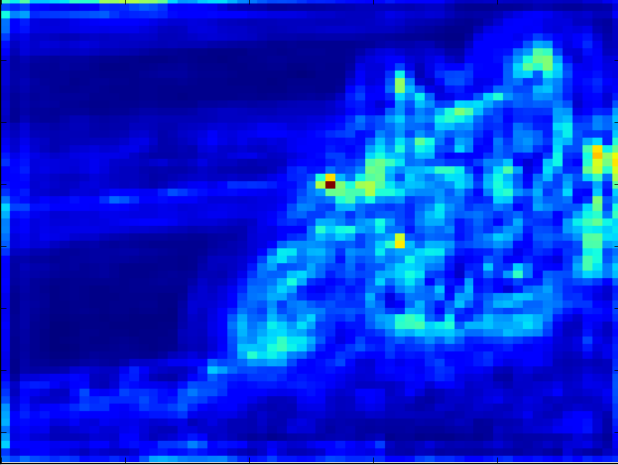} &
\includegraphics[width=.08\textwidth]{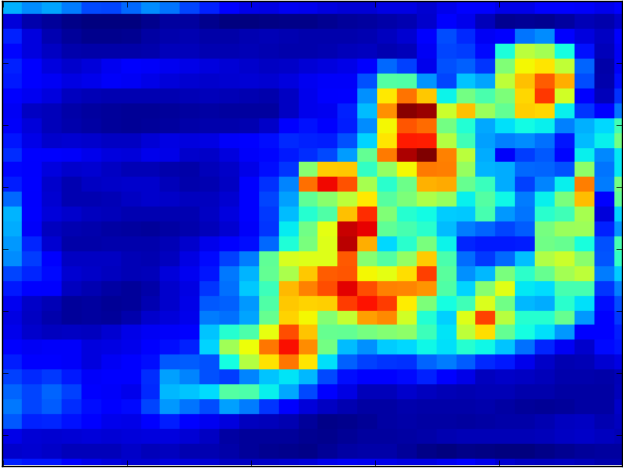} &
\includegraphics[width=.08\textwidth]{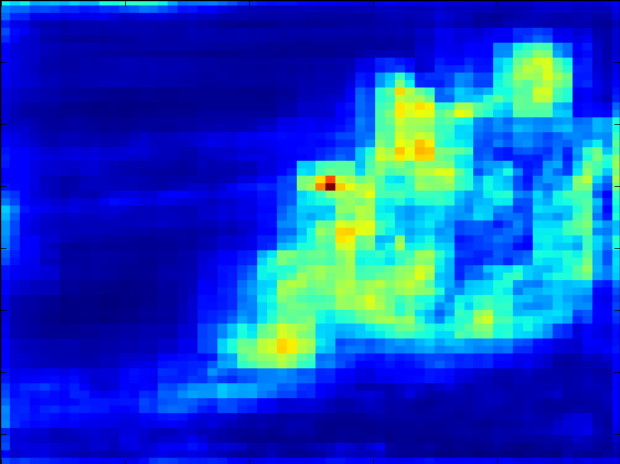} &
\includegraphics[width=.08\textwidth]{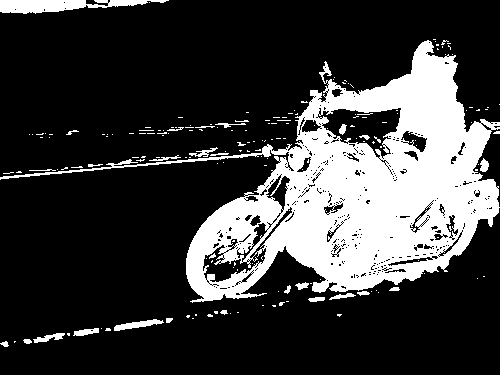} &
\includegraphics[width=.08\textwidth]{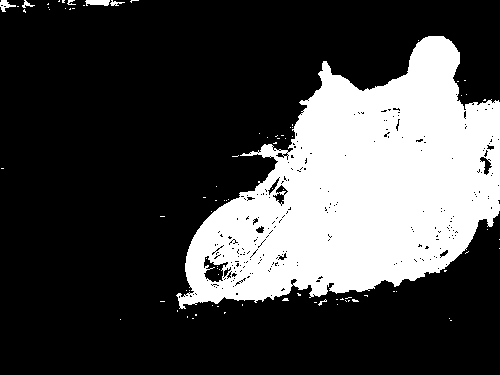} &
\includegraphics[width=.08\textwidth]{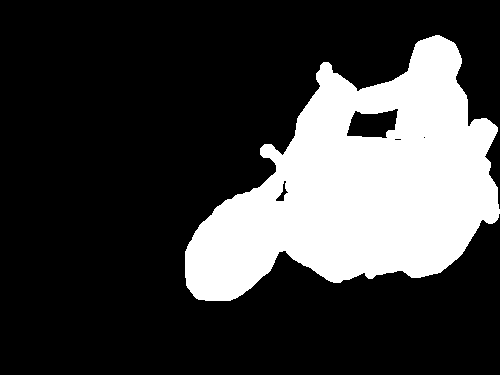} \\

\includegraphics[width=.08\textwidth]{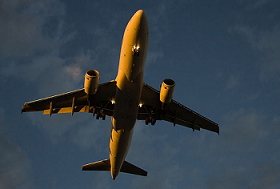} &
\includegraphics[width=.08\textwidth]{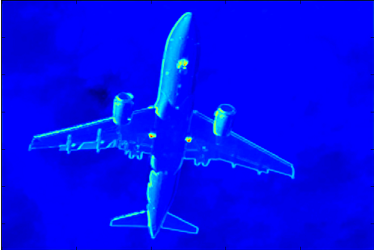} &
\includegraphics[width=.08\textwidth]{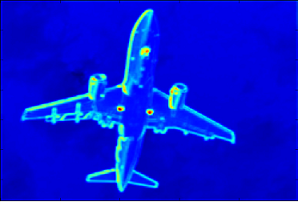} &
\includegraphics[width=.08\textwidth]{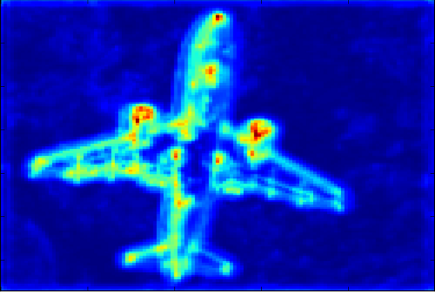} &
\includegraphics[width=.08\textwidth]{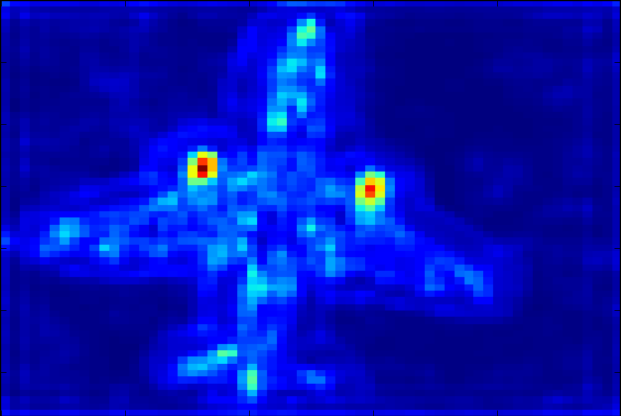} &
\includegraphics[width=.08\textwidth]{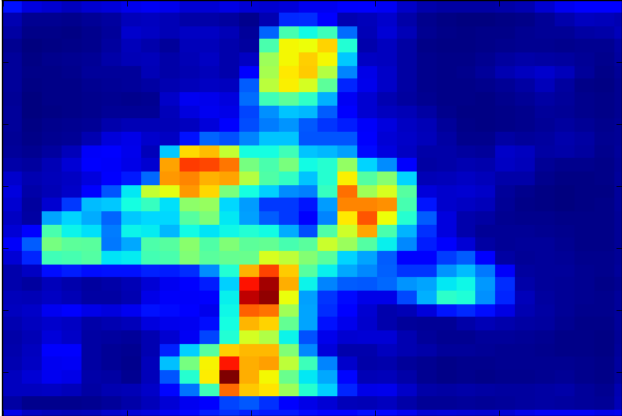} &
\includegraphics[width=.08\textwidth]{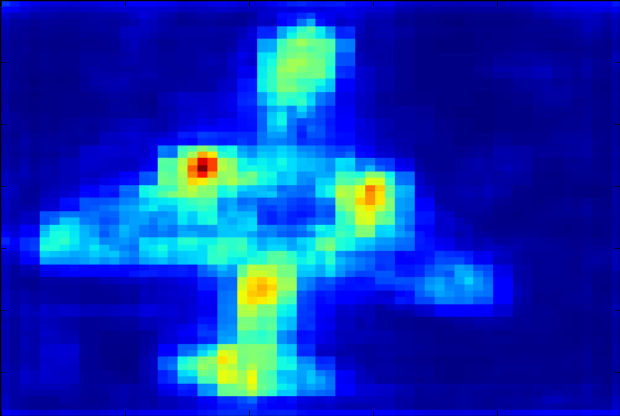} &
\includegraphics[width=.08\textwidth]{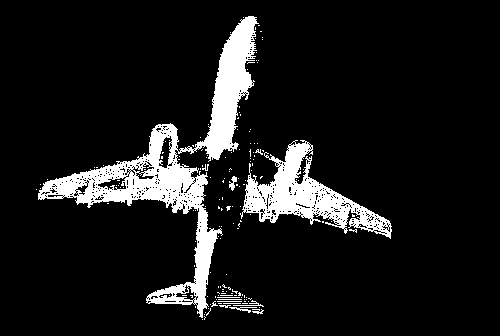} &
\includegraphics[width=.08\textwidth]{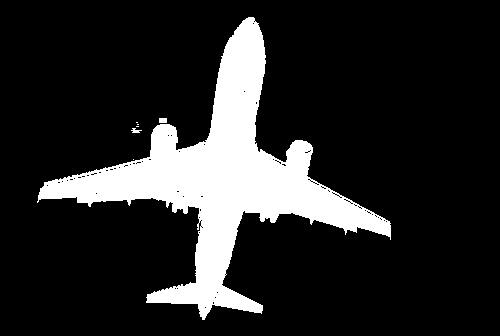} &
\includegraphics[width=.08\textwidth]{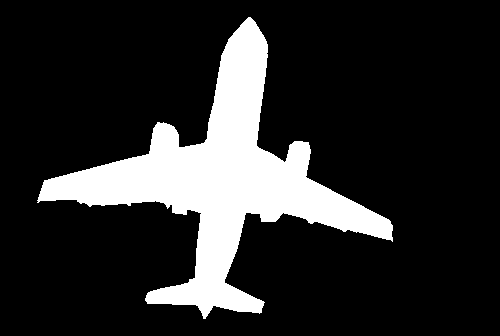} \\

\includegraphics[width=.08\textwidth]{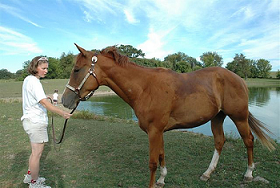} &
\includegraphics[width=.08\textwidth]{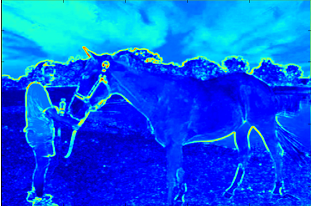} &
\includegraphics[width=.08\textwidth]{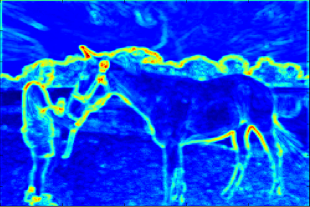} &
\includegraphics[width=.08\textwidth]{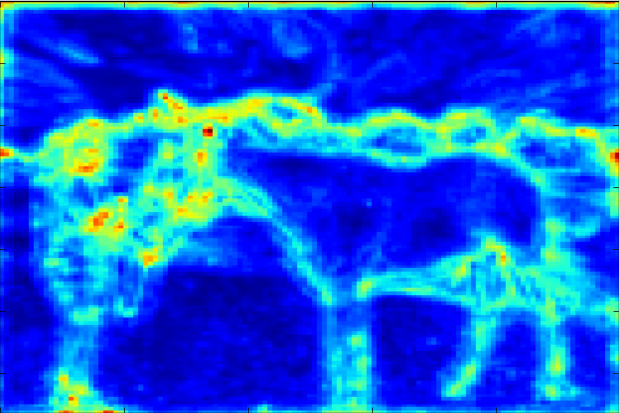} &
\includegraphics[width=.08\textwidth]{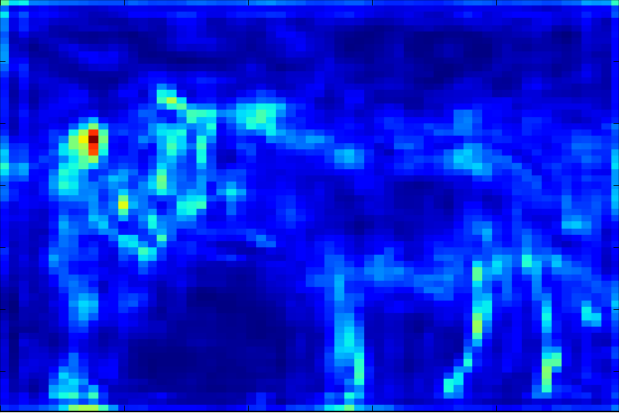} &
\includegraphics[width=.08\textwidth]{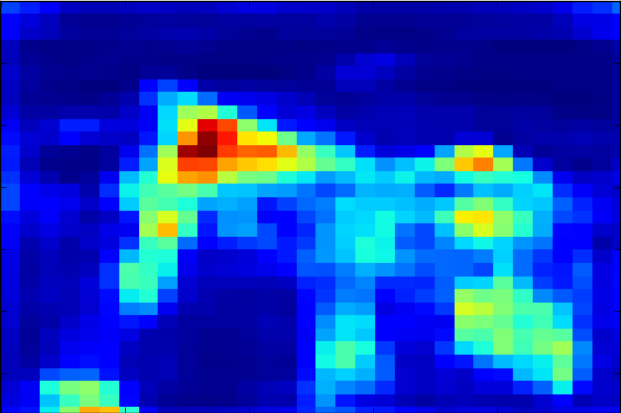} &
\includegraphics[width=.08\textwidth]{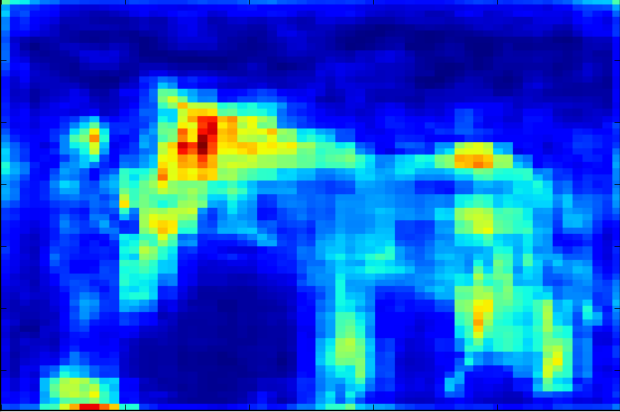} &
\includegraphics[width=.08\textwidth]{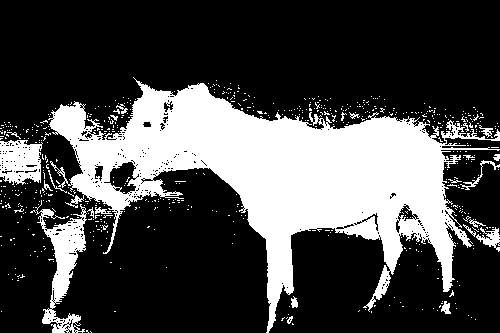} &
\includegraphics[width=.08\textwidth]{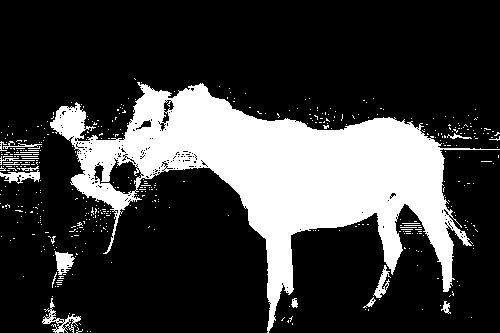} &
\includegraphics[width=.08\textwidth]{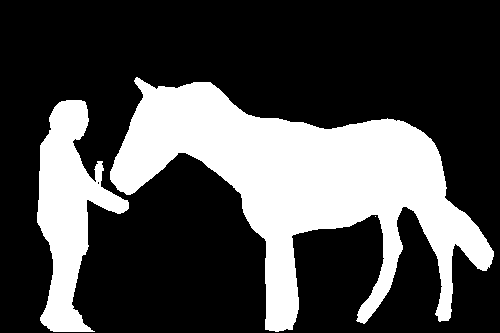} \\

\end{tabular}
\caption{{\bf Built-in foreground/background mask.} From left to right, we show the input image, the activations of the first, second, third, fourth, and fifth convolutional layers, the results of our fusion strategy, and the final mask after CRF smoothing without and with higher order followed by the ground-truth mask. Note that "Fusion" constitutes the unary potential of the dense CRF used to obtain "Our mask".}
\label{fig:fusion}
\end{figure*}

\subsubsection{Foreground/Background Masks}
\label{sec:fb}
Let us first consider the case of foreground/background masks. In practice, as discussed in more detail in Section~\ref{sec:network}, we make use of an architecture based on the VGG-16 network~\cite{VGG16}, whose weights were trained on ImageNet for the task of object recognition, converted into a fully-convolutional network. If, to recognize objects, the network has learned to focus on the object themselves, it should produce high activation values on the objects and on their parts. To evaluate this, we studied the activations of the different hidden layers of our initial network. 

More specifically, we passed each image forward through the network and visualized each activation by computing the mean over the channels after resizing the activation map to the input image size. Perhaps unsurprisingly, this lead to the following observations, illustrated in Fig.~\ref{fig:fusion}. The first two convolutional layers of the VGG network extract image edges. As we move deeper in the network, the convolutional layers extract higher-level features. In particular, the third convolutional layer fires up on prototypical object shapes. The fourth and fifth layers indicate the location of complete objects and of their most discriminative parts. Note that a similar study was performed in the different context of edge detection~\cite{deepedge}, with similar conclusions. 

Based on these observations, we propose to make use of the fourth and fifth layers to produce an initial foreground/background mask estimate. To this end, we first convert these two layers from 3D tensors (\(512\times W\times H\)) to 2D matrices (\(W\times H\)) via an average pooling operation over the 512 channels. We then fuse the two resulting matrices by simple elementwise summation, and scale the resulting values between 0 and 1. The resulting $W\times H$ map can be thought of as a pixelwise foreground probability, which we denote by $P_f$ in the remainder of the paper. Fig.~\ref{fig:fusion} illustrates the results of this method on a few images from PASCAL VOC 2012. Note that, while the resulting scores indeed accurately indicate the location of the foreground objects, this initial mask remains noisy. This will be addressed in Section~\ref{sec:CRF} by encouraging smoothness via a CRF.

Our foreground/background masks can be thought of as a form of objectness measure. While objectness has been used previously for weakly-supervised semantic segmentation (MCG and BING in~\cite{fromimgtopixel,augment_feedback,learnTosegwithimagelevelanno}, and the generic objectness~\cite{objectness} in~\cite{whatsthepoint}), the benefits of our approach are twofold. First, we extract this information directly from the same network that will be used for semantic segmentation, which prevents us from having to rely on an external method. Second, as opposed to BING and MCG, we require neither object bounding boxes, nor object segments to train our method. Finally, as shown in our experiments, our method yields much more accurate object localization than the techniques in~\cite{objectness} and~\cite{MCG}, which typically only provide a rough outline of the objects.

\subsubsection{Multi-class Masks}
\label{sec:mc}

The main drawback of the foreground/background masks discussed above is that they are not class-specific. The network we used to extract these masks has not been fine-tuned with the desired classes, and thus the activations only provide information about the location of generic foreground objects. Here, we address this limitation by making use of a class-specific localization network~\cite{loc} in conjunction with our foreground/background masks.

The main idea behind the localization network of~\cite{loc} is to generate a Class Activation Map (CAM) for each specific object category, or, in other words, a heat map indicating the location of the regions that are useful for the network to recognize a specific category. This is achieved by making use of the global average pooling strategy of~\cite{network_in_network}, and importantly, without using any bounding box, or pixel-level annotations.

In our case, as discussed in Section~\ref{sec:loc_network}, our starting point is a fully-convolutional version of the VGG-16 network. Just before the final output layer (the cross entropy loss layer for multi class categorization), we perform global average pooling on the convolutional feature maps and use the resulting features as input to a fully-connected layer that produces class scores. Specifically, let $f_k(x,y)$ denote the activation of unit $k$ at spatial location $(x,y)$ in the last convolutional layer, and $F^k = \sum_{x,y}{f_k(x,y)}$ the result of global average pooling for unit $k$. Then, the predicted score for a given class $c$ can be written as $S_c = \sum_k{w_k^c F^k}$, where $w_k^c$ is the weight corresponding to class $c$ for unit $k$. In essence, $w_k^c$ indicates the importance of unit $k$ for class $c$.

To generate a CAM, one can thus rely on these weights. In particular, these weights are used in a linear combination of the activations of the units in the last convolutional layer. This lets us express a CAM for class $c$ as
\begin{equation}
M_c(x,y) = \sum_k{w_k^c f_k(x,y)}\;.
\end{equation}
Ultimately, $M_c(x,y)$ directly indicates how important the observations at spatial grid $(x,y)$ are to classify the input image as belonging to class $c$.

\begin{figure}[!t]
\centering
\tiny
\begin{tabular}{c c c }
Image & $CAM_{horse}$ & $CAM_{person}$  \\

\includegraphics[width=.1\textwidth]{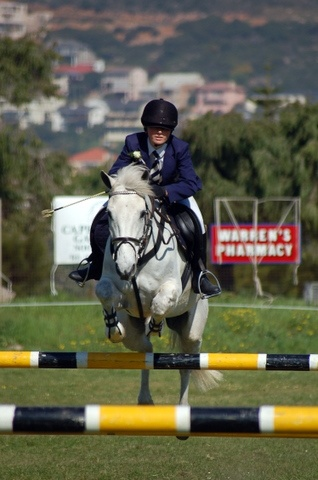} &
\includegraphics[width=.1\textwidth]{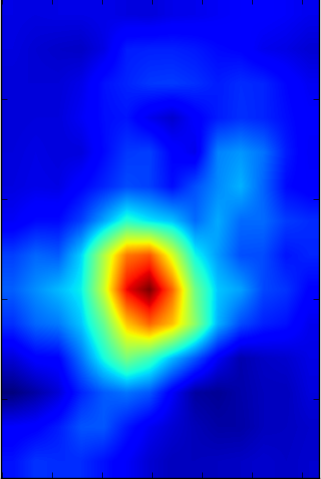} &
\includegraphics[width=.1\textwidth]{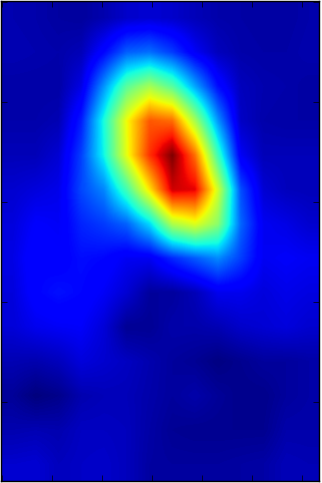}  \\
Image & $CAM_{motorbike}$ & $CAM_{person}$  \\
\includegraphics[width=.1\textwidth]{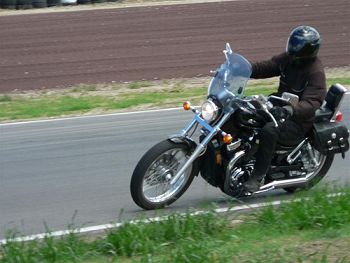} &
\includegraphics[width=.1\textwidth]{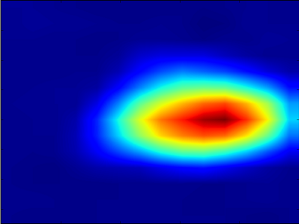} &
\includegraphics[width=.1\textwidth]{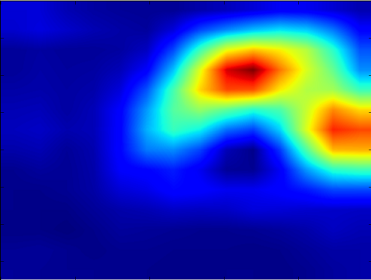} \\
Image & $CAM_{horse}$ & $CAM_{person}$  \\
\includegraphics[width=.1\textwidth]{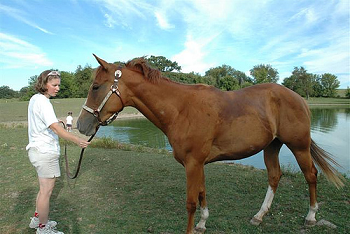} &
\includegraphics[width=.1\textwidth]{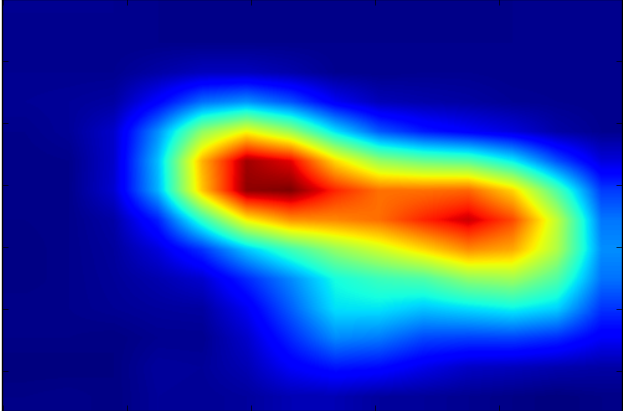} &
\includegraphics[width=.1\textwidth]{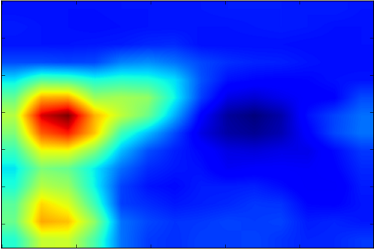} \\

\end{tabular}
\caption{CAM for each class obtained by the localization network.} 
\label{fig:loc_only}
\end{figure}

As can be seen in Fig.~\ref{fig:loc_only}, the resulting CAMs suffer from two main drawbacks. First, they only roughly match the shape of the object, yielding inaccurate localization of the object's boundary. Second, they typically only focus on the discriminative parts of the objects, which is sufficient for object recognition, but not for segmentation. To overcome these limitations, we propose to combine these CAMs with our foreground/background masks, to obtain more accurate and more complete multi-class masks.

To this end, and as suggested in~\cite{loc}, we first generate binary masks from each $M_c$ by setting to 1 the values that are above 20\% of the maximum value in each $M_c$, and to 0 the other ones. Let us denote by $B_c$ the resulting binary mask for class $c$. From these binary masks and the foreground/background probabilities $P_f$ obtained by fusing the activations of the fourth and fifth convolutional layers, we form a new multi-class mask, which, for each class $c$, is defined as a map
\begin{equation}
Q_c = P_{f}\odot B_c\;,
\end{equation}
where we think of each map as a matrix, and where $\odot$ indicates the Hadamard (elementwise) product. This, in essence, can be thought of as a class-specific truncated version of $P_f$, where the truncation masks are obtained from the $M_c$s, with a permissive threshold of 20\% to avoid cutting out too many regions.

\begin{figure}[!t]
\centering
\tiny
\begin{tabular}{c c c }
Image & Ground-Truth &\\
\includegraphics[width=.1\textwidth]{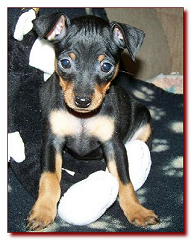} &
\includegraphics[width=.1\textwidth]{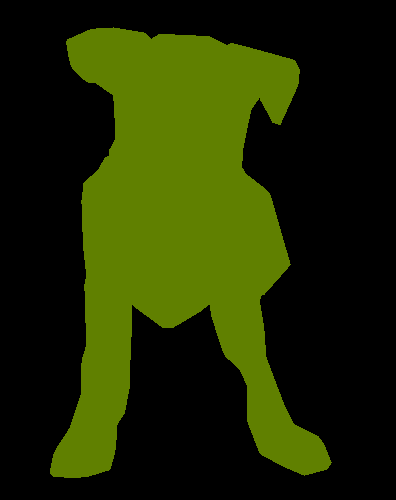} & \\
Fusion & Localization & $Q_c$+Localization \\
\includegraphics[width=.1\textwidth]{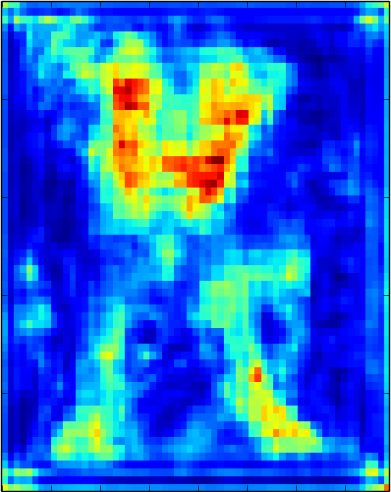} &
\includegraphics[width=.1\textwidth]{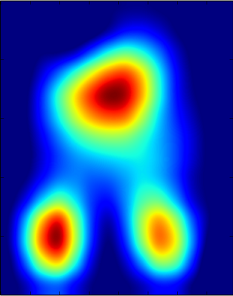} &
\includegraphics[width=.1\textwidth]{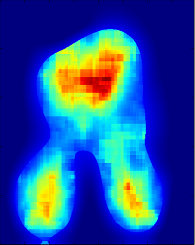} \\
fg/bg mask w/ & multi-class mask w/  & multi-class mask w/\\
 Fusion unaries & localization unaries & $Q_c$+localization unaries \\
 \includegraphics[width=.1\textwidth]{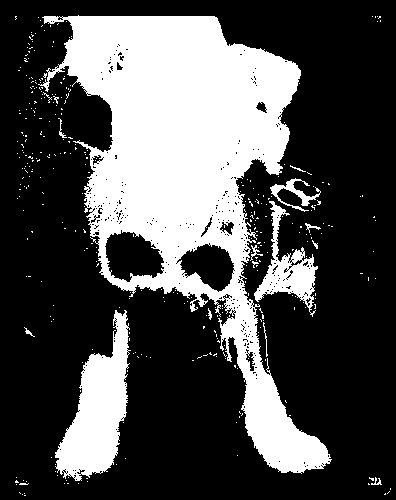} &
 \includegraphics[width=.1\textwidth]{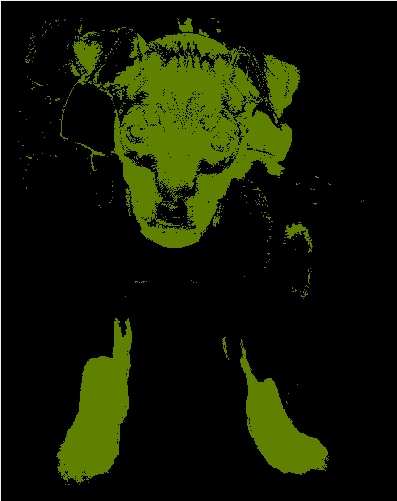} &  \includegraphics[width=.1\textwidth]{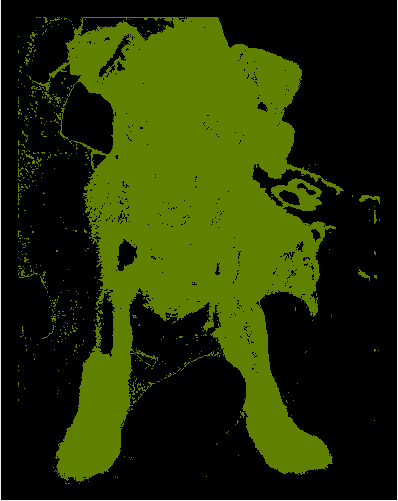} \\
\end{tabular}
\caption{Effect of adding localization information to our Fusion map ($Q_c$).} 
\label{fig:loc_fusion}
\end{figure}

To obtain our final multi-class masks, we combine these class-specific truncated fusion maps with the original CAMs. To this end, we make use of a linear combination, which yields, for each class $c$, the final map
\begin{equation}
P_c = \alpha \cdot Q_c + (1-\alpha) \cdot M_c\;,
\end{equation}
where, in practice, we set $\alpha = 0.5$, and which is normalized to obtain a probability. The resulting probabilities are compared to the fusion-based ones and to the CAMs in Fig.~\ref{fig:loc_fusion}. Note that the final maps preserve the more accurate boundary information and the better object coverage of the fusion-based ones, while removing their noise, thanks to the CAMs.

At this point, we have probability maps for each foreground class $c$, but not for the background class. To generate such a background map, we simply use the probabilities of the locations that have not been considered as foreground classes in $M_c$. To this end, we define
\begin{equation}
M_{0} = 1 - \frac{1}{C} \sum_c{M_c}\;,
\end{equation}
which, in turn, lets us define the background map as
\begin{equation}
P_0 = \alpha \cdot (1-P_{f})  + (1-\alpha) \cdot M_{0}.\;
\end{equation}

While better than both our foreground/background masks and the CAMs, our multi-class masks remain noisy. To address this, in the next section, we propose to make use of a fully-connected CRF with higher-order terms.

\subsubsection{Smoothing the Masks with a Dense CRF}
\label{sec:CRF}

To smooth out initial noisy masks, we make use of a fully-connected CRF with higher-order terms. Note that, while we consider the general, multi-class case, the formalism discussed below applies to both our foreground/background masks and to our multi-class masks.

Specifically, let ${\bf x} = \{x_i\}_{i=1}^{W\cdot H}$ be the set of random variables, where $x_i$ encodes the label of pixel $i$, i.e., either one of the foreground classes or background. We encode the joint distribution over all pixels with a Gibbs energy of the form
\begin{align}
E({\bf x} = {\bf X}) = -\sum_i \theta_i(x_i = X_i)  \nonumber \\+ \sum_{i}\sum_{j>i}\theta_{ij}(x_i=X_i, x_j=X_j) \label{eq:gibbs} \\ + \sum_{{\bf x}_s\in \mathcal{R}} \theta_s({\bf x}_s = {\bf X}_s)\;, \nonumber
\end{align}
where $\theta_i$ is a unary potential defining the cost of assigning label $X_i$ to pixel $i$, and the second and third terms encode pairwise and higher-order potentials, respectively, with $\mathcal{R}$ a set of regions. 

The unary potential is obtained directly from the probability maps introduced in either Section~\ref{sec:fb} or~\ref{sec:mc} as
\begin{align}
\theta_i(x_i = X_i) = -\log\left(\frac{\exp\left(P(x_i=X_i)\right)}{\sum_{l=1}^C{\exp\left(P(x_i = l)\right)}}\right)\;,
\label{eq:unary}
\end{align}
where $P$ can be either $P_f$ or $P_c$. 

The pairwise potential $\theta_{ij}$  encodes the compatibility of a joint label assignment for two pixels. Following~\cite{DenseCRF}, we define this pairwise term as a contrast-sensitive Potts model using two Gaussian kernels encoding color similarity and spatial smoothness. Such a model penalizes two pixels at relatively close spatial locations and with similar appearance to be assigned different labels.

\begin{figure}[!t]
\centering
\tiny
\begin{tabular}{c c c c  c}
Image & Ground-Truth & DCRF  & Crisp Segments~\cite{Crisp} & DCRF \\
& & &  & + higher order\\
\includegraphics[width=.09\textwidth]{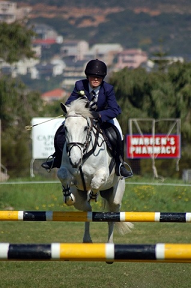} &
\includegraphics[width=.09\textwidth]{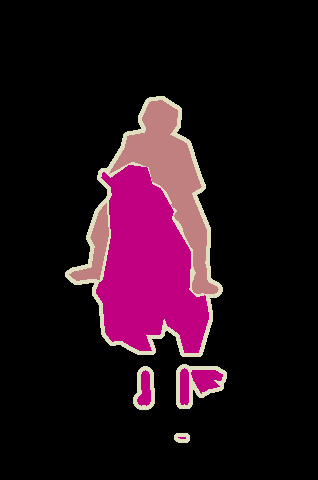} &
\includegraphics[width=.09\textwidth]{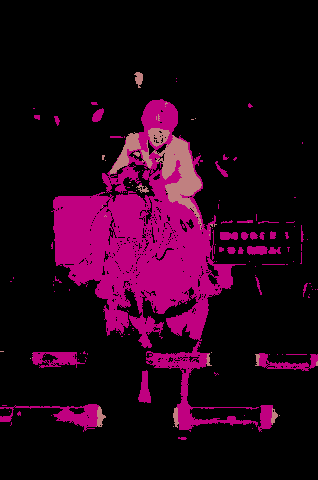} &
\includegraphics[width=.09\textwidth]{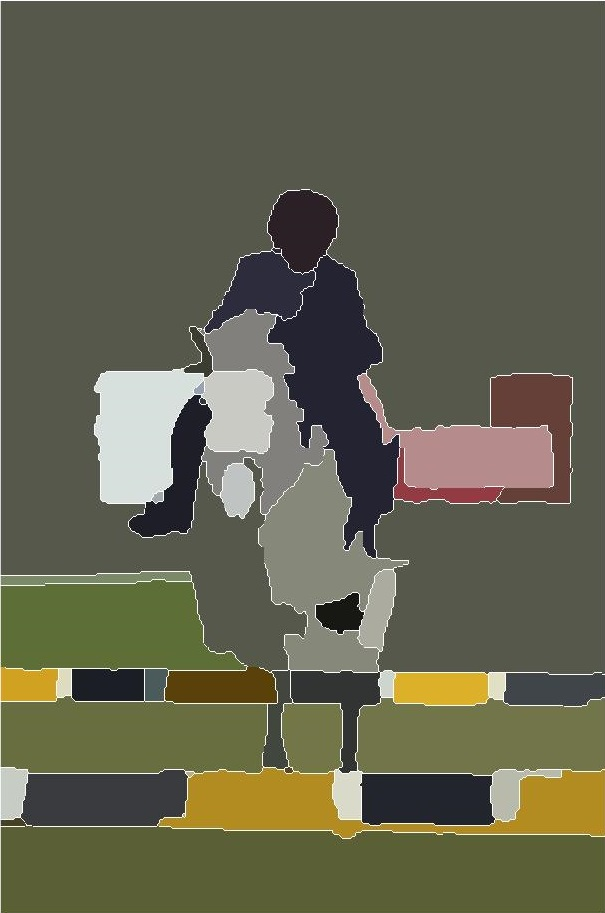} &
\includegraphics[width=.09\textwidth]{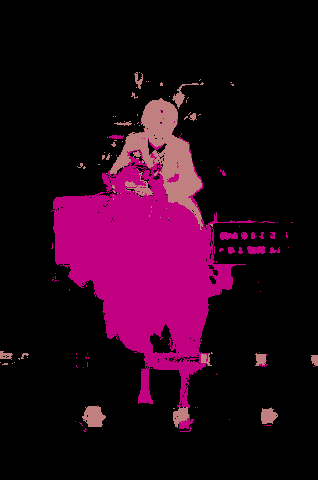} \\

\includegraphics[width=.09\textwidth]{localization/2007_005951_img.png} &
\includegraphics[width=.09\textwidth]{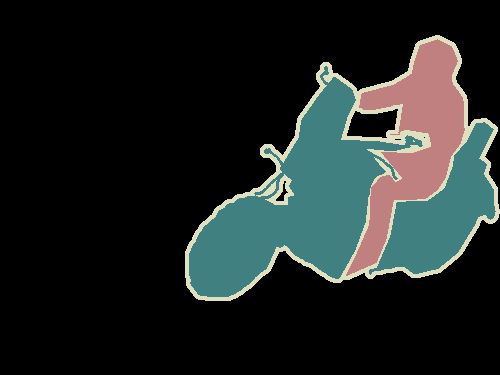} &
\includegraphics[width=.09\textwidth]{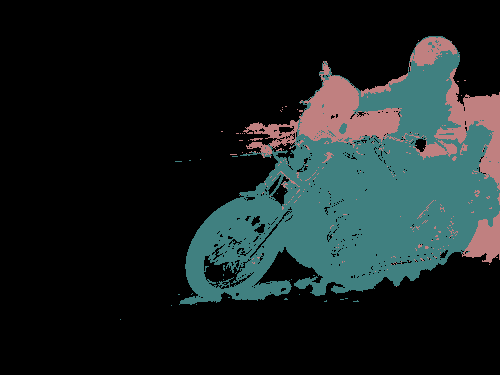} &
\includegraphics[width=.09\textwidth]{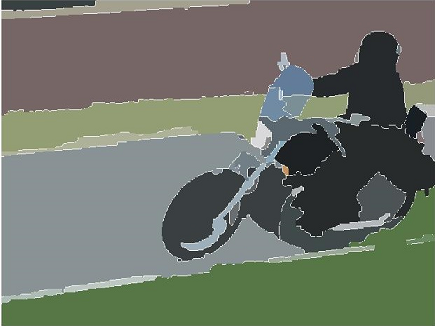} &
\includegraphics[width=.09\textwidth]{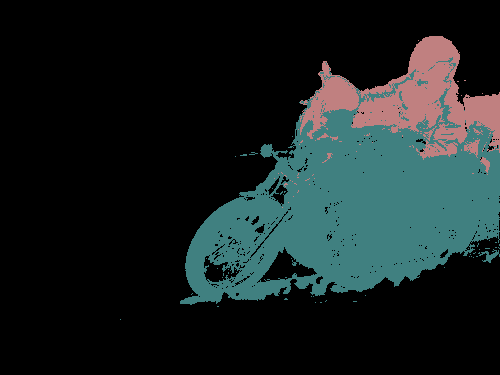} \\

\includegraphics[width=.09\textwidth]{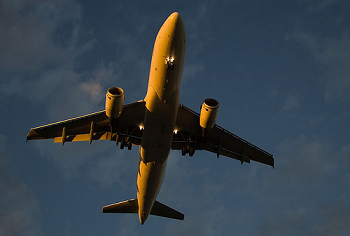} &
\includegraphics[width=.09\textwidth]{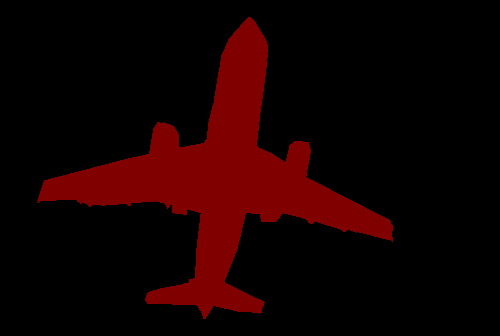} &
\includegraphics[width=.09\textwidth]{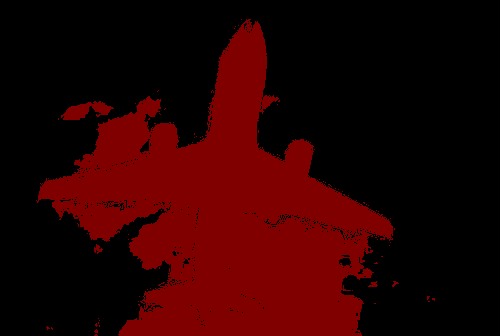} &
\includegraphics[width=.09\textwidth]{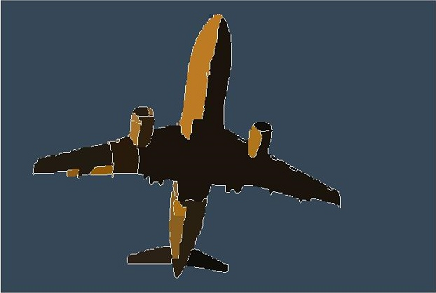} &
\includegraphics[width=.09\textwidth]{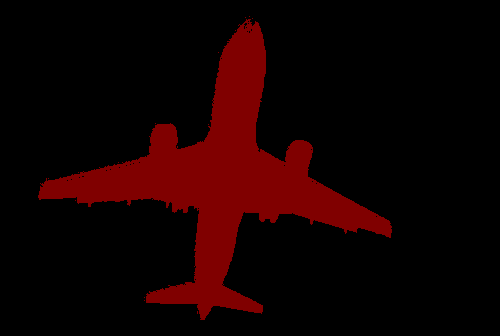} \\

\includegraphics[width=.09\textwidth]{localization/2008_000142_img.png} &
\includegraphics[width=.09\textwidth]{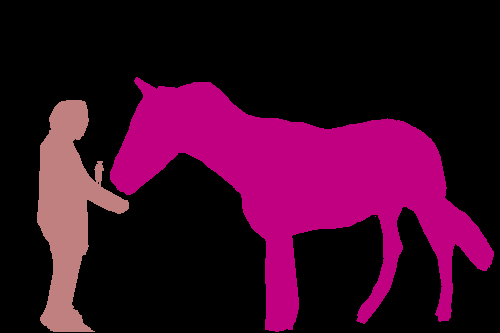} &
\includegraphics[width=.09\textwidth]{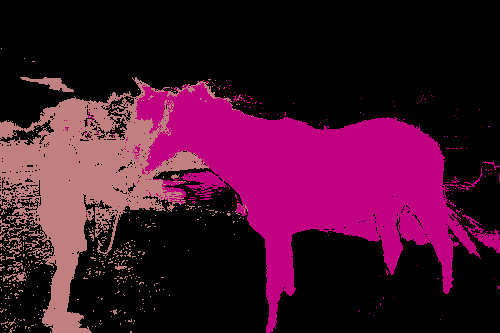} &
\includegraphics[width=.09\textwidth]{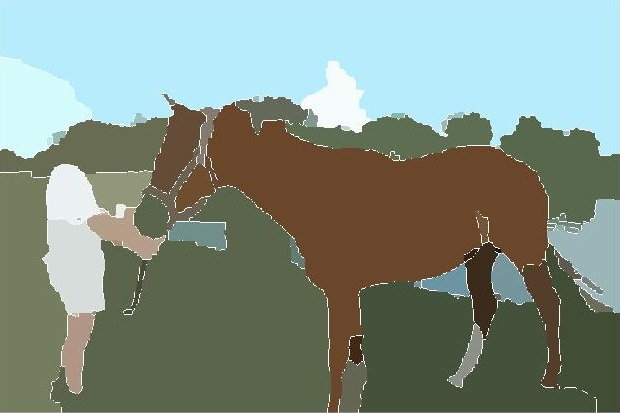} &
\includegraphics[width=.09\textwidth]{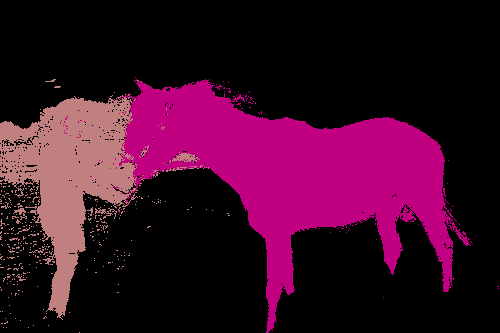} \\

\end{tabular}
\caption{Effect of using higher-order potentials using regions obtained by the crisp boundary detection method of~\cite{Crisp}.} 
\label{fig:higher_order}
\end{figure}

For the higher-order terms, we make use of a $P^n$-Potts model encouraging all the pixels in one region to be assigned the same label. To define the regions, we propose to make use of the crisp boundary detection algorithm of~\cite{Crisp}. This algorithm aims at detecting the boundaries between semantically different objects visible in the scene. It is based on a simple underlying principle: pixels belonging to the same object exhibit higher statistical dependencies than pixels belonging to different objects. This method is unsupervised and can adapt to each input image independently. As illustrated in Fig.~\ref{fig:higher_order}, the resulting crisp boundaries can be thought of as defining semantically coherent regions, which are thus very well-suited to our goal. For each region ${\bf x}_s$, we then define the cost of the higher-order term as
\begin{align}
\theta_s({\bf x}_s = l)=-\log\left(\frac{\sum_{x_i\in {\bf x}_s}P(x_i=l)}{N_s}\right),
\label{eq:ho_potential}
\end{align}
if all the pixels are assigned the same label $l$, and a maximum cost otherwise. Here, $N_s$ indicates the number of pixels in region $s$.

By using Gaussian pairwise potentials and $P^n$-Potts higher-order ones, we can make use of the inference strategy of~\cite{vineet_higherorder}, which relies on the filtering-based mean-field method of~\cite{DenseCRF}. In Figs.~\ref{fig:fusion}--\ref{fig:higher_order}, we show the effect of CRF smoothing on our masks with and without higher-order terms.

\setlength{\tabcolsep}{0.2em}
\setlength{\tabcolsep}{0.5em}
\subsection{Weakly-Supervised Learning}
We now introduce our learning algorithm for weakly-supervised semantic segmentation. We first introduce a simple loss based on image tags only, and then show how we can incorporate our two different types of masks in this framework.

Intuitively, given image tags, one would like to encourage the image pixels to be labeled as one of the classes that are observed in the image, while preventing them to be assigned to unobserved classes. Note that this assumes that the tags cover all the classes depicted in the image. This assumption, however, is commonly employed in weakly-supervised semantic segmentation~\cite{whatsthepoint,MIL,fromimgtopixel}. Formally, given an input image $I$, let $\mathcal{L}$ be the set of classes that are present in the image (including background) and $\bar{\mathcal{L}}$ the set of classes that are absent. Furthermore, let us denote by $s_{i,j}^k(\theta)$ the score produced by our network with parameters $\theta$ for the pixel at location ($i,j$) and for class $k$, $0\leq k < N$. Note that, in general, we will omit the explicit dependency of the variables on the network parameters. Finally, let $S_{i,j}^k$ be the probability of class $k$ obtained after a softmax layer, i.e., 

\begin{equation}
S_{i,j}^k = \frac{\exp(s_{i,j}^k)}{\sum_{c=1}^N\exp(s_{i,j}^c)}\;.
\end{equation}

Encoding the above-mentioned intuition can then simply be achieved by designing a loss of the form
\begin{equation}
L_{weak} = -\frac{1}{|\mathcal{L}|}\sum_{k\in \mathcal{L}} \log{S^k} - \frac{1}{|\bar{\mathcal{L}}|}\sum_{k\in \bar{\mathcal{L}}}\log(1-S^k) \;,
\label{eq:loss_mil}
\end{equation}
where $S^k$ represents a candidate score for each class in the image. In short, the first term in Eq.~\ref{eq:loss_mil} expresses the fact that the present classes should be in the image, while the second term penalizes the pixels that have high probabilities for the absent classes. In practice, instead of computing $S^k$ as the maximum probability (as previously used in~\cite{MIL,whatsthepoint}) for class $k$ over all pixels in the image, we make use of the convex Log-Sum-Exp (LSE) approximation of the maximum (as previously used in~\cite{fromimgtopixel}), which can be written as
\begin{equation}
S^k = \frac{1}{r}\log\left[\frac{1}{|I|}\sum_{i,j \in I}\exp(rS_{i,j}^k)\right]\;,
\label{eq:LSE}
\end{equation}
where $|I|$ denotes the total number of pixels in the image and $r$ is a parameter allowing this function to behave in a range between the maximum and the average. In practice, following~\cite{fromimgtopixel}, we set $r$ to 5.

The loss in Eq.~\ref{eq:loss_mil} does not rely on any notion of foreground and background. As a consequence, minimizing it will typically yield poor object localization accuracy. To overcome this issue, we propose to make use of our built-in priors introduced in Sections~\ref{sec:fb} and~\ref{sec:mc}. Below, we start with the foreground/background case, and then turn to the multi-class scenario.

\subsubsection{Incorporating Foreground/Background Masks}
\label{sec:learning1}

When only a foreground/background probability is available, we cannot directly reason at the level of specific classes. Instead, we rely on this mask to encourage all pixels labeled as one of the object tags to lie within a foreground region, while the other pixels should belong to the background. 

To this end, let $M_{i,j}$ denote the mask value at pixel $(i,j)$, i.e., $M_{i,j} = 1$ if pixel $(i,j)$ belongs to the foreground and 0 otherwise. We can then re-write our loss as
\begin{align}
L_{mask} = -\frac{1}{|\mathcal{L}|-1}\sum_{k\in \mathcal{L}, k \neq 0}\log(S^k) -\log(S^0) \nonumber \\ - \frac{1}{|\bar{\mathcal{L}}|\cdot|I|}\sum_{i,j\in I,\;k\in \bar{\mathcal{L}}}\log(1-S_{i,j}^k)\;,
\label{eq:loss_mil_mask}
\end{align}
where
\begin{equation}
S^k = \frac{1}{r}\log\left[\frac{1}{|M|}\sum_{i,j | M_{i,j}=1}\exp(rS_{i,j}^k)\right]\;,
\end{equation}
and
\begin{equation}
S^0 = \frac{1}{r}\log\left[\frac{1}{|\bar{M}|}\sum_{i,j | M_{i,j}=0}\exp(rS_{i,j}^0)\right]\;,
\end{equation}
with $|M|$ and $|\bar{M}|$ the number of foreground and background pixels, respectively. $S^k$ computes an approximate maximum probability for the present class $k$ over all pixels in the foreground mask. Similarly, $S^0$ denotes an approximate maximum probability for the background class over all pixels outside the foreground mask. In short, the loss of Eq.~\ref{eq:loss_mil_mask} favors present classes to appear in the foreground mask, while pixels predicted as background should be assigned to the background class and no pixels should take on an absent label.

To learn the parameters of our network, we follow a standard back-propagation strategy to search for the parameters $\theta$ that minimize the loss in Eq.~\ref{eq:loss_mil_mask}. In particular, the network is fine-tuned using stochastic gradient descent (SGD) with momentum to update the weights by a linear combination of the negative gradient and the previous weight update. At inference time, given the test image, the network performs a dense prediction. We optionally apply a fully-connected CRF with higher-order terms similar to the one discussed above to smooth the segmentation.

\subsubsection{Incorporating Multi-class Masks}
\label{sec:learning2}
In the presence of multi-class masks, we can then reason about the specific classes that are observed in the input image. In this scenario, we would like to encourage the pixels set to 1 in one particular class mask corresponding to one input tag to be assigned the label of this class. Enforcing this strongly, e.g., by considering the maximum score over all pixels in a mask, would unfortunately be sensitive to noise in the mask, as further discussed in our experiments. Instead, here, we propose to again make use of the LSE to have a softer penalty.

Specifically, let $M_k$ be the mask corresponding to image tag, i.e., class label, $k$. We propose to take into account our multi-class masks by re-writing our loss function as
\begin{equation}
L_{weak} = -\frac{1}{|\mathcal{L}|}\sum_{k\in \mathcal{L}}\log(S^k) - \frac{1}{|\bar{\mathcal{L}}|\cdot|I|}\sum_{i,j\in I,\;k\in \bar{\mathcal{L}}}\log(1-S_{i,j}^k)\;,
\label{eq:loss_mil_multimask}
\end{equation}
where
\begin{equation}
S^k = \frac{1}{r}\log\left[\frac{1}{|M_k|}\sum_{i,j | M_{k}=1}\exp(rS_{i,j}^k)\right]\;.
\end{equation}
In other words, this loss encourages, for each present class $k$, including the background class, the pixels belonging to the corresponding mask to be assigned label $k$, while penalizing the pixels that take on an absent label. We use the same learning strategy as in the foreground/background case to minimize this. Furthermore, as before, during inference, the network provides a dense labeling for an input test image, without requiring any tag, and this labeling can optionally be smoothed via CRF inference.

\section{Experimental Results}
In this section, we first describe the datasets used for our experiments, and then provide details about our learning and inference procedures. We then compare our method with foreground/background masks and with multi-class ones to the state-of-the-art weakly supervised semantic segmentation algorithms. Finally, we provide a thorough evaluation of the effect of the different components of our approach.

\subsection{Datasets}

\textbf{PASCAL VOC 2012}. In our experiments, we made use of the standard Pascal VOC 2012 dataset~\cite{pascal}, which serves as a benchmark in most weakly-supervised semantic segmentation papers~\cite{whatsthepoint,weaklyandsemi,MIL,fromimgtopixel,CCNN}. This dataset contains $21$ classes, and 10,582 training images (the VOC 2012 training set and the additional data annotated by~\cite{augmenteddata}), 1,449 validation images and 1,456 test images. The image tags were obtained from the pixel-level annotations by simply listing the classes observed in each image. As in~\cite{whatsthepoint,weaklyandsemi,fromimgtopixel,CCNN}, we report results on both the validation and the test set.

\textbf{YouTube Objects}. This dataset (YTO)~\cite{yto} contains videos collected from YouTube by querying for the names of 10 object classes of the PASCAL VOC dataset. It contains between 9 and 24 videos per class. For our experiments, we uniformly extracted around 2200 frames per class to obtain a total of 22k frames out of 700k available in the dataset. For evaluation we use the subset of images with pixel-level annotations provided by~\cite{mask-yto}. Note that there is no overlap between this subset and the shots from which we extracted the training data.

\textbf{Microsoft COCO}. For MS COCO~\cite{coco}, we made use of 80k training samples with only image-level tags to train our network and 40k validation samples to evaluate the performance of our method. The MS COCO annotations were designed for instance level labeling. As such, some pixels in the images can be assigned multiple labels. For example, a pixel can belong to both Fork and Dining Table. To evaluate our results for semantic segmentation, we obtained a unique ground-truth label per pixel by using the label of the smallest object, that is, fork in the example above.

Note that Sections~\ref{sec:comp} and~\ref{sec:ablation} focus on the PASCAL VOC dataset, which is the one commonly used for weakly-supervised semantic segmentation. The results for YTO and MS COCO, which demonstrate the generality of our method, are provided in Section~\ref{sec:coco}.

\subsection{Implementation Details}
\label{sec:impl}
\subsubsection{Semantic Segmentation Networks}
\label{sec:network}
As most recent weakly-supervised semantic segmentation algorithms~\cite{whatsthepoint,weaklyandsemi,MIL,fromimgtopixel,CCNN,SEC}, our architecture is based on the VGG-16 network~\cite{VGG16}, whose weights were trained on ImageNet for the task of object recognition. Following the fully-convolutional approach~\cite{FCN}, all fully-connected layers are converted to convolutional layers, and the final classifier replaced with a \(1\times 1\) convolution layer with $N$ channels, where $N$ represents the number of classes of the problem. We use two different versions of this fully-convolutional network. When utilizing foreground/background masks, inspired by~\cite{DeepLab}, we used a stride of 8 and a relatively small receptive field of 128 pixels, which has proven effective in practice for weakly-supervised semantic segmentation~\cite{weaklyandsemi}. By contrast, when using multi-class masks, inspired by~\cite{DeepLab} again, we found that using a larger field of view improves the results. We therefore employed a kernel size of $3\times3$ in the convolutional layer corresponding to the first fully connected layer of VGG-16 and an input stride of 12, resulting in a receptive field size of 224. We also reduced the number of filters from 4096 to 1024 to allow for faster training~\cite{DeepLab}. With both types of masks, at the end of the network, we added a deconvolution layer to up-sample the output of the network to the size of the input image. In short, the network takes an image of size $W\times H$ as input and generates an $N\times W\times H$ output encoding a score for each pixel and for each class.

For both types of masks, the parameters of the network were found using stochastic gradient descent with a learning rate of $10^{-4}$ for the first 40k iterations and $10^{-5}$ for the next 20k iterations, a momentum of $0.9$, a weight decay of $0.0005$, and mini-batches of size 1. Similarly to recent weakly-supervised segmentation methods~\cite{fromimgtopixel,CCNN,MIL,whatsthepoint,weaklyandsemi}, the network weights were initialized with those of a network pre-trained for a 1000-way classification task on the ILSVRC 2012 dataset~\cite{ILSVRC}. Hence, for the last convolutional layer, we used the weights corresponding to the 20 classes shared by Pascal VOC and ILSVRC. For the background class, we initialized the weights with zero-mean Gaussian noise with a standard deviation of 0.1.

At inference time, given the test image, but no tags, the network generates a dense prediction as a complete semantic segmentation map. We used C++ and Python (Caffe framework~\cite{caffe}) for our implementation. As other methods~\cite{CCNN,weaklyandsemi,SEC}, we further optionally apply a dense CRF to refine this initial segmentation. As mentioned in Section~\ref{sec:CRF}, we added higher-order potentials to the dense pairwise CRF.

\begin{table*}[t!]
\renewcommand{\arraystretch}{1.2}
\centering
\caption{Per class IOU on the PASCAL VOC 2012 validation set for methods trained using image tags. }
\label{tab:pascal_tags}
\scalebox{0.75}
{
\hspace{-0.4cm}\begin{tabular}{| l|c c c c c c c c c c c c c c c c c c c c c|c |} 
\hline
Method & \rotatebox[origin=c]{90}{bg} & \rotatebox[origin=c]{90}{aero} & \rotatebox[origin=c]{90}{bike} & \rotatebox[origin=c]{90}{bird} & \rotatebox[origin=c]{90}{boat} & \rotatebox[origin=c]{90}{bottle} & \rotatebox[origin=c]{90}{bus} & \rotatebox[origin=c]{90}{car} & \rotatebox[origin=c]{90}{cat} & \rotatebox[origin=c]{90}{chair} & \rotatebox[origin=c]{90}{cow} & \rotatebox[origin=c]{90}{table} & \rotatebox[origin=c]{90}{dog} & \rotatebox[origin=c]{90}{horse} & \rotatebox[origin=c]{90}{mbike} & \rotatebox[origin=c]{90}{person} & \rotatebox[origin=c]{90}{plant} & \rotatebox[origin=c]{90}{sheep}& \rotatebox[origin=c]{90}{sofa} & \rotatebox[origin=c]{90}{train} & \rotatebox[origin=c]{90}{tv} & \rotatebox[origin=c]{0}{mIOU} \\
 \hline

MIL(Tag)~\cite{fromimgtopixel} & 37.0  & 10.4  & 12.4 &  10.8 & 5.3 & 5.7 & 25.2 & 21.1 & 25.15 & 4.8 & 21.5 & 8.6 & 29.1 & 25.1 & 23.6 & 25.5 & 12.0 & 28.4 & 8.9 & 22.0 & 11.6 & 17.8\\

MIL(Tag) w/ILP~\cite{fromimgtopixel} & 73.2 & 25.4 & 18.2 & 22.7 & 21.5 & 28.6 & 39.5 & 44.7 & 46.6 & 11.9 & 40.4 & 11.8 & 45.6 & 40.1 & 35.5 & 35.2 & 20.8 & 41.7 & 17.0 & 34.7 & 30.4 & 32.6\\

MIL(Tag) w/ILP+sspxl~\cite{fromimgtopixel} & 77.2 & 37.3 & 18.4 & 25.4 & 28.2 & 31.9 & 41.6 & 48.1 & 50.7 & 12.7 & 45.7 & 14.6 & 50.9 & 44.1 & 39.2 & 37.9 & 28.3 & 44.0 & 19.6 & 37.6 & 35.0 & 36.6\\

What's the point(Tag) W/Obj~\cite{whatsthepoint}& 78.8& 41.6& 19.8& 38.7& 33.0& 17.2& 33.8& 38.8& 45.0& 10.4& 35.2& 12.6& 42.3& 34.3& 33.2& 22.7& 18.6& 40.1& 14.9& 37.7& 28.1& 32.2\\

EM-Fixed(Tag)+CRF~\cite{weaklyandsemi} & - & - & - & - & - & - & - & - & - & - & - & - & - & - & - & - & - & - & - & - & - & 20.8 \\

EM-Adapt(Tag)+CRF~\cite{weaklyandsemi} & - & - & - & - & - & - & - & - & - & - & - & - & - & - & - & - & - & - & - & - & - & 38.2 \\

CCNN(Tag)~\cite{CCNN} & 66.3 & 24.6 & 17.2 & 24.3 & 19.5 & 34.4 & 45.6 & 44.3 & 44.7 & 14.4 & 33.8 & 21.4 & 40.8 & 31.6 & 42.8 & 39.1 & 28.8 & 33.2 & 21.5 & 37.4 & 34.4 & 33.3\\

CCNN(Tag)+CRF~\cite{CCNN} & 68.5 & 25.5 & 18.0 & 25.4 & 20.2 & 36.3 & 46.8 & 47.1 & 48.0 & 15.8 & 37.9 & 21.0 & 44.5 & 34.5 & 46.2 & 40.7 & 30.4 & 36.3 & 22.2 & 38.8 & 36.9 & 35.3\\

SEC+CRF~\cite{SEC} & 82.2 & 61.7 & 26.0 & 60.4 & 25.6 & 45.6 & 70.9 & 63.2 & 72.2 & 20.9 & 52.9 & 30.6 & 62.8 & 56.8 & 63.5 & 57.1 & 32.2 & 60.6 & 32.3 & 44.8 & 42.3 & 50.7\\
DCSM+CRF~\cite{distinct}&76.7 &45.1& 24.6 &40.8 &23.0 &34.8 &61.0 &51.9& 52.4& 15.5& 45.9& 32.7& 54.9& 48.6& 57.4& 51.8& 38.2& 55.4& 32.2& 42.6& 39.6& 44.1\\

fg/bg masks+CRF~\cite{OurECCV} & 79.2 & 60.1 & 20.4 & 50.7 & 41.2 & 46.3 & 62.6 & 49.2 & 62.3 & 13.3 & 49.7 & 38.1 & 58.4 & 49.0 & 57.0 & 48.2 & 27.8 & 55.1 & 29.6 & 54.6 & 26.6 & 46.6\\
 \hline

Ours fg/bg masks+CRF &
   82.0 &   68.0  & 26.9  & 66.5 & 34.1 &   47.4 &   57.3 &   51.7 &   72.2 &   14.5 &   50.6 &   26.6 &   65.3 &   55.9 &   58.7 &   25.8 & 29.7 &   62.5 &   27.9 &   54.1 &   30.0 & 48.0 \\
 Ours multi-class masks+CRF & 82.2 &   59.5 &   27.4 &   66.7 &   25.2 &   44.1 &   71.1 &   55.1 &   71.9 &   19.7 &   52.3 &   36.7 &   65.6 &   59.4 &   62.8 &   55.3 &   32.3 & 65.5 &   34.3 &   43.4 &   38.8 & 50.9\\
  \hline
\end{tabular}
}
\end{table*}

\begin{table*}[!t]
\renewcommand{\arraystretch}{1.2}
\centering

\caption{Per class IOU on the PASCAL VOC 2012 test set for methods trained using image tags.}
\label{tab:pascal_test}
\scalebox{0.75}
{
\hspace{-0.2cm}\begin{tabular}{| l|c c c c c c c c c c c c c c c c c c c c c|c |} 
\hline
Method & \rotatebox[origin=c]{90}{bg} & \rotatebox[origin=c]{90}{aero} & \rotatebox[origin=c]{90}{bike} & \rotatebox[origin=c]{90}{bird} & \rotatebox[origin=c]{90}{boat} & \rotatebox[origin=c]{90}{bottle} & \rotatebox[origin=c]{90}{bus} & \rotatebox[origin=c]{90}{car} & \rotatebox[origin=c]{90}{cat} & \rotatebox[origin=c]{90}{chair} & \rotatebox[origin=c]{90}{cow} & \rotatebox[origin=c]{90}{table} & \rotatebox[origin=c]{90}{dog} & \rotatebox[origin=c]{90}{horse} & \rotatebox[origin=c]{90}{mbike} & \rotatebox[origin=c]{90}{person} & \rotatebox[origin=c]{90}{plant} & \rotatebox[origin=c]{90}{sheep}& \rotatebox[origin=c]{90}{sofa} & \rotatebox[origin=c]{90}{train} & \rotatebox[origin=c]{90}{tv} & \rotatebox[origin=c]{0}{mIOU}\\
 \hline
 CCNN (tags)+CRF~\cite{CCNN} & - & 24.2 & 19.9 & 26.3 & 18.6 & 38.1 &  51.7 &  42.9 &  48.2 &  15.6  & 37.2  & 18.3 &  43.0  & 38.2  & 52.2  & 40.0  & 33.8 - & 36.0  & 21.6  & 33.4 &  38.3  & 35.6\\
MIL-FCN~\cite{fromimgtopixel} & - & - & - & - & - & - & - & - & - & - & - & - & - & - & - & - & - & - & - & - & - & 25.7\\
MIL-sppxl~\cite{fromimgtopixel} & 74.7 & 38.8 & 19.8 & 27.5 & 21.7 & 32.8 & 40.0 & 50.1 & 47.1 & 7.2 & 44.8 & 15.8 & 49.4 & 47.3 & 36.6 & 36.4 & 24.3 & 44.5 & 21.0 & 31.5 & 41.3 & 35.8\\
MIL-obj~\cite{fromimgtopixel} & 76.2 & 42.8 & 20.9 & 29.6 & 25.9 & 38.5 & 40.6 & 51.7 & 49.0 & 9.1 & 43.5 & 16.2 & 50.1 & 46.0 & 35.8 & 38.0 & 22.1 & 44.5 & 22.4 & 30.8 & 43.0 & 37.0\\
EM-Adapt+CRF~\cite{weaklyandsemi} & 76.3 & 37.1 & 21.9 & 41.6 & 26.1 & 38.5 & 50.8 & 44.9 & 48.9 & 16.7 & 40.8 & 29.4 & 47.1 & 45.8 & 54.8 & 28.2 & 30.0 & 44.0 & 29.2 & 34.3 & 46.0 & 39.6\\
SEC+CRF~\cite{SEC} & 83.0 & 55.6 & 27.4 & 61.1 & 22.9 & 52.4 & 70.2 & 58.8 & 70.0 & 22.1 & 54.3 &  27.9 & 67.4 & 59.4 & 70.7 & 59.0 & 38.7 & 58.6 & 38.1 & 37.6 & 45.2 & 51.5\\
DCSM+CRF~\cite{distinct}&78.1 &43.8& 26.3& 49.8& 19.5& 40.3& 61.6& 53.9& 52.7& 13.7& 47.3& 34.8& 50.3& 48.9& 69.0& 49.7& 38.4& 57.1& 34.0& 38.0& 40.0& 45.1\\
fg/bg masks+CRF~\cite{OurECCV} &   80.3 & 57.5 & 24.1 & 66.9 & 31.7 & 43.0 & 67.5 & 48.6 & 56.7 & 12.6 & 50.9 & 42.6 & 59.4 & 52.9 & 65.0 & 44.8 & 41.3 & 51.1 & 33.7 & 44.4 & 33.2 & 48.0\\
 \hline
Ours fg/bg masks+CRF & 83.4 & 65.4& 29.0& 68.5& 33.4& 51.6& 58.4& 53.5& 68.3& 15.7& 54.1& 30.2& 66.9& 57.9& 66.0& 23.7& 39.6& 61.6& 29.7& 51.9& 31.8& 49.6\\

Ours multi-class masks+CRF & 83.5 & 60.8 & 29.8 & 66.6 & 23.2 & 52.1 & 69.3 & 53.8 & 70.4 & 19.1 & 56.8 & 40.1 & 71.0 & 59.7 & 71.4 & 54.9 & 33.9 & 71.2 & 40.5 & 35.4 & 41.9 & 52.6\\  
\hline
\end{tabular}
}
\end{table*}

\begin{table}
\renewcommand{\arraystretch}{1.1}
\centering
\caption{Mean IOU on the PASCAL VOC validation and test sets for other methods trained with higher level of supervision or additional training data. Note that, while our approach requires no additional supervision or training data, its accuracy is comparable to or higher than that of other methods.}
\label{tab:pascal_sup}
\begin{tabular}{l|c|c}
 Methods & mIoU(val) & mIOU(test)\\
\hline
\cite{fromimgtopixel}: MIL(Tag) w/ILP+bbox & 37.8 & 37.0 \\
\cite{fromimgtopixel}: MIL(Tag) w/ILP+seg & 42.0 & 40.6\\
\cite{learnTosegwithimagelevelanno}: SN-B+MCG seg & 41.9 & 43.2\\
\cite{whatsthepoint}: 1Point  & 35.1 & -\\ 
\cite{whatsthepoint}: Objectness+1Point  & 42.7& -\\
\cite{whatsthepoint}: Objectness+1Point(GT) & 46.1 & -\\
\cite{whatsthepoint}: Objectness+AllPoints (weighted)  & 43.4 &- \\
\cite{whatsthepoint}: Objectness+1 squiggle per class  & 49.1 &- \\
\cite{CCNN}: Random Crops+CRF  & 36.4 & -\\
\cite{CCNN}: Size Info.+CRF  & 42.4 & 45.1\\
\cite{STC}: STC + CRF + additional train data  & 49.8 &51.2 \\
\cite{learnTosegwithimagelevelanno} SN-B+MCG seg & 41.9 & 43.2\\
\cite{OurECCV}: CheckMask procedure+CRF & 51.5 & 52.9\\
\cite{augment_feedback}: Augmented feedback+MCG+CRF & 54.3 & 55.5\\
\hline
Ours (fg/bg masks)+CRF & 48.0 & 49.6\\
Ours (multi-class masks)+CRF & 50.9 & 52.6 \\
\end{tabular}
\end{table}

\subsubsection{Localization Network}
\label{sec:loc_network}
For the localization network, we followed the approach introduced in~\cite{loc}. Specifically, the architecture of the network was again derived from the VGG-16 architecture~\cite{VGG16}, pre-trained for the task of object recognition on ImageNet. We then substituted the last two fully-connected layers, fc6 and fc7, with randomly initialized convolutional layers. The output of the last convolutional layer acts as input to a global average pooling layer followed by a fully-connected prediction layer corresponding to the number of foreground classes of interest (20 for PASCAL VOC).  The network was fine-tuned for object recognition on the training set of the PASCAL VOC 2012 dataset with a cross entropy loss. To this end, we used images of size of $224\times 224$ as input, and mini-batches of size 15. The other optimization parameters were set to the same values as for the semantic segmentation network. 

Note that we could in principle also fine-tune the VGG-16 network used to generate our foreground/background masks for object recognition on the target dataset (e.g., PASCAL VOC). In practice, however, we observed that this did not improve the quality of our masks. 

\begin{figure*}[!t]
\centering
\tiny
\begin{tabular}{cccccccc}
 Image & baseline & fg/bg mask & fg/bg mask+HO & multi-class mask/Loc & multi-class mask/Loc &multi-class mask/Loc& GT  \\
  &  &  &  & & +Fusion &+Fusion+HO &   \\
\includegraphics[width=.09\textwidth]{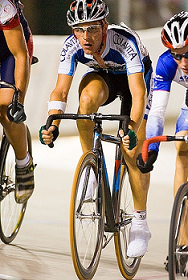} &
\includegraphics[width=.09\textwidth]{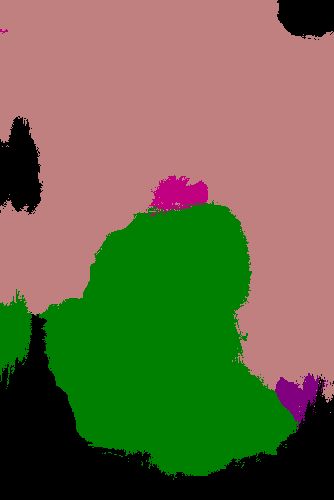} &
\includegraphics[width=.09\textwidth]{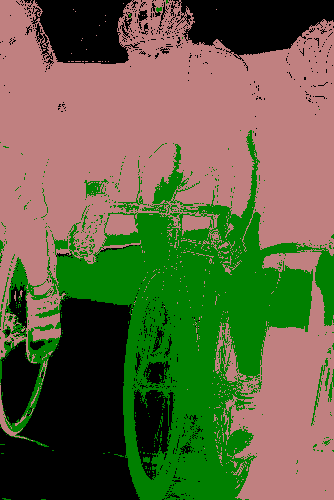} &
\includegraphics[width=.09\textwidth]{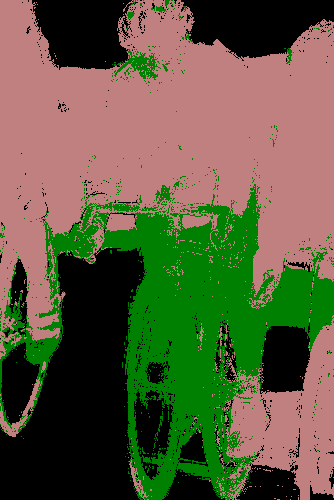} &
\includegraphics[width=.09\textwidth]{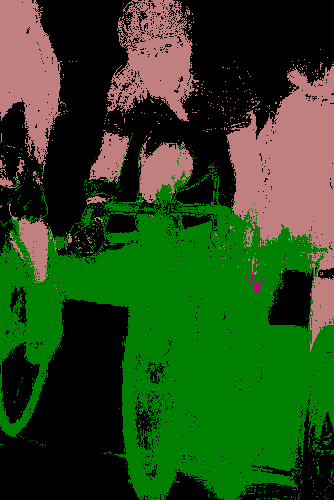} &
\includegraphics[width=.09\textwidth]{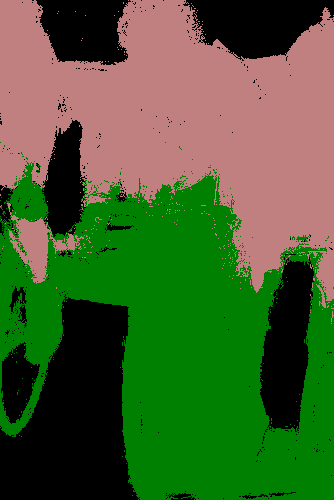} &
\includegraphics[width=.09\textwidth]{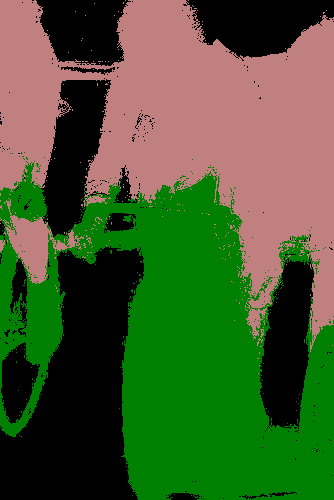} & \includegraphics[width=.09\textwidth]{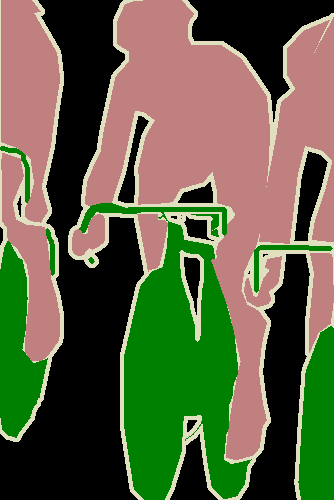}\\

\includegraphics[width=.09\textwidth]{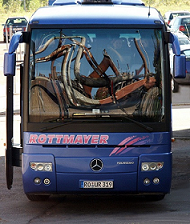} &
\includegraphics[width=.09\textwidth]{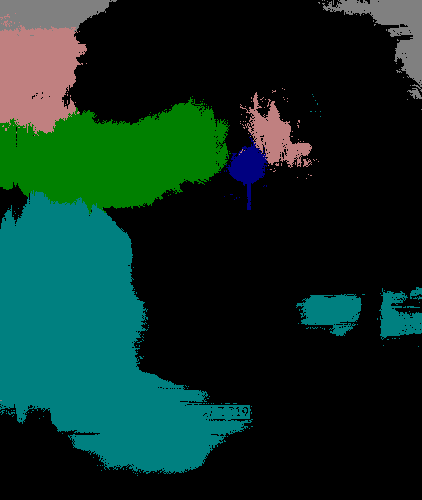} &
\includegraphics[width=.09\textwidth]{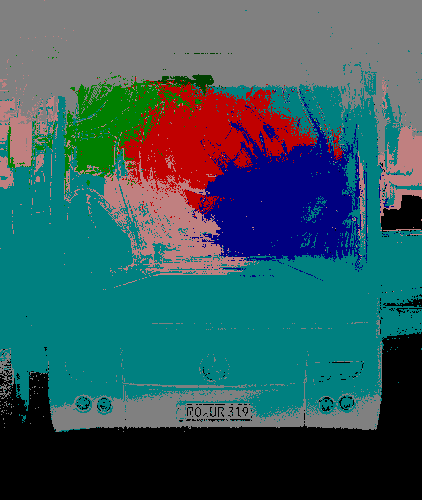} &
\includegraphics[width=.09\textwidth]{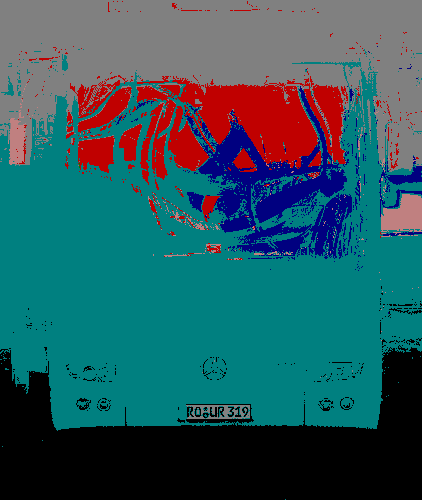} &
\includegraphics[width=.09\textwidth]{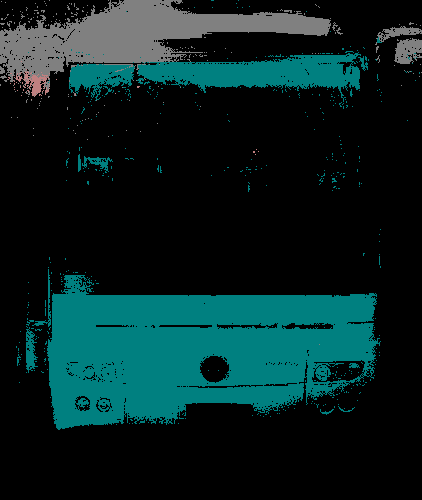} &
\includegraphics[width=.09\textwidth]{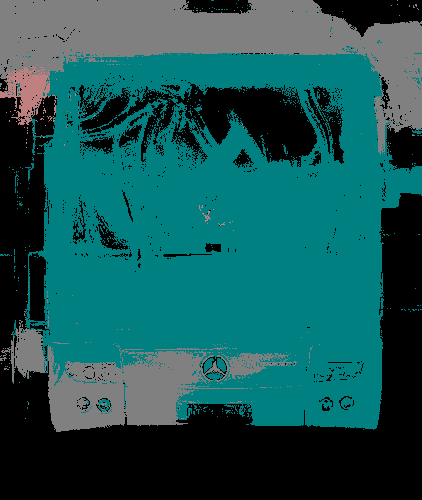} &
\includegraphics[width=.09\textwidth]{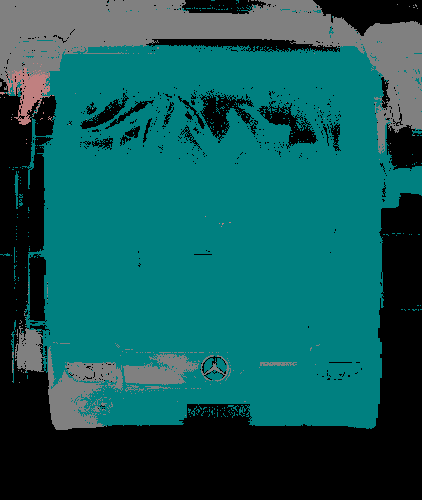} & \includegraphics[width=.09\textwidth]{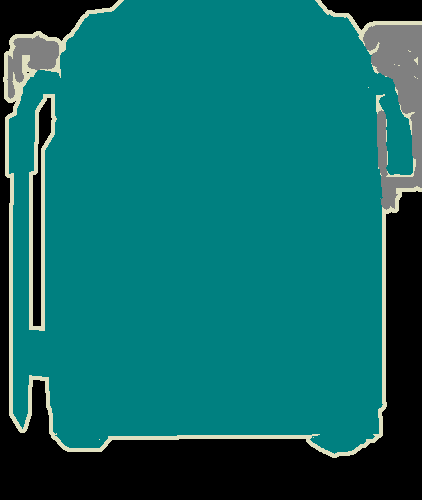}\\

\includegraphics[width=.09\textwidth]{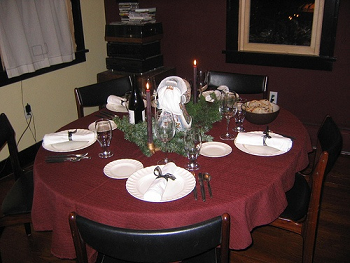} &
\includegraphics[width=.09\textwidth]{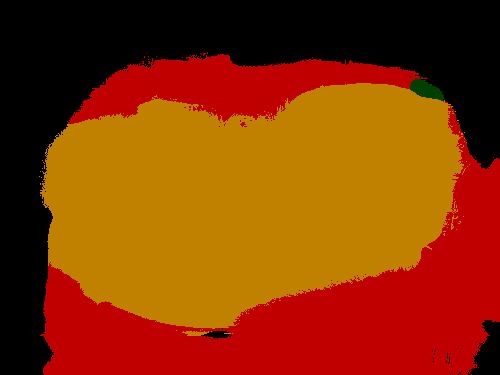} &
\includegraphics[width=.09\textwidth]{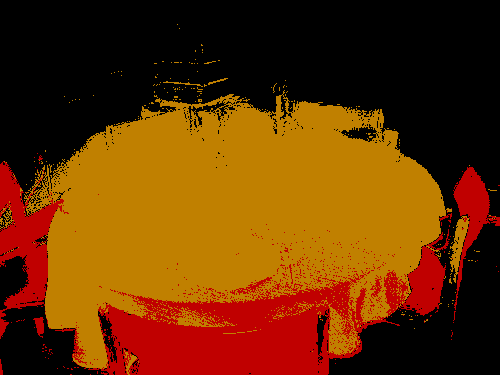} &
\includegraphics[width=.09\textwidth]{Results/2007_006241_Bmask.png} &
\includegraphics[width=.09\textwidth]{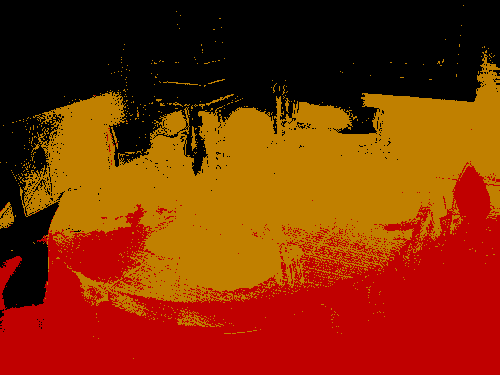} &
\includegraphics[width=.09\textwidth]{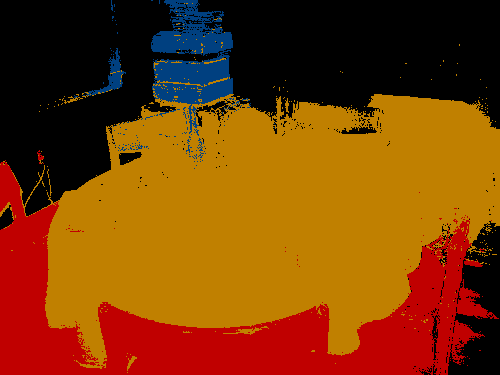} &
\includegraphics[width=.09\textwidth]{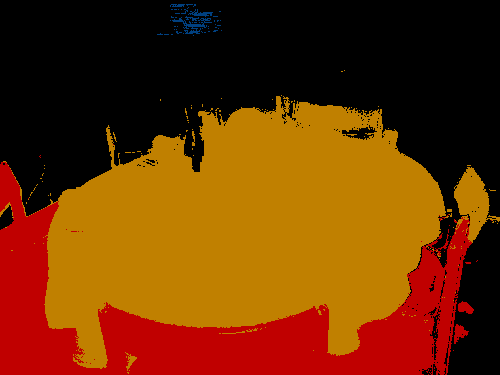} & \includegraphics[width=.09\textwidth]{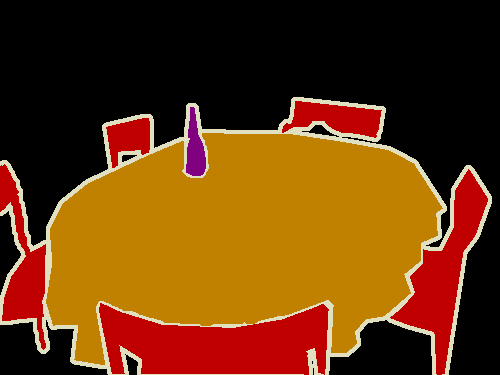}\\

\includegraphics[width=.09\textwidth]{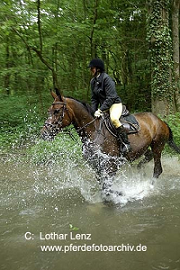} &
\includegraphics[width=.09\textwidth]{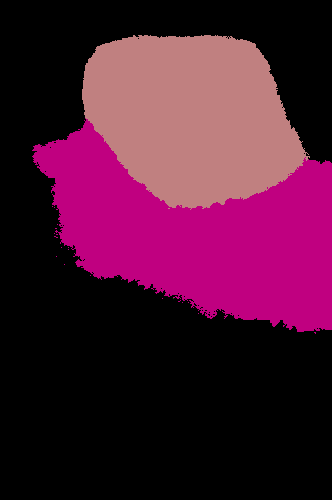} &
\includegraphics[width=.09\textwidth]{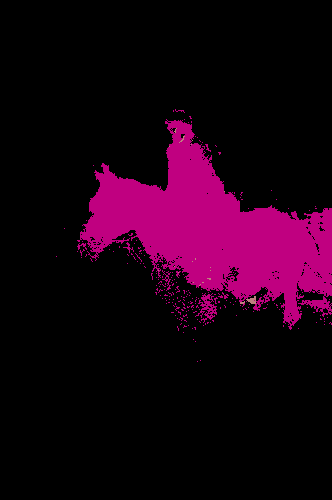} &
\includegraphics[width=.09\textwidth]{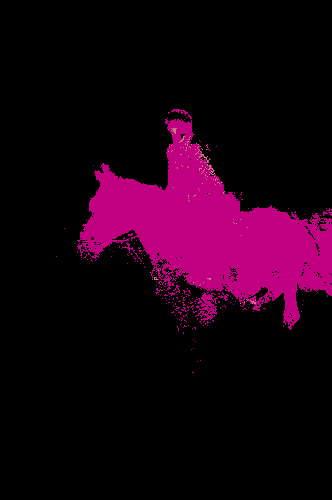} &
\includegraphics[width=.09\textwidth]{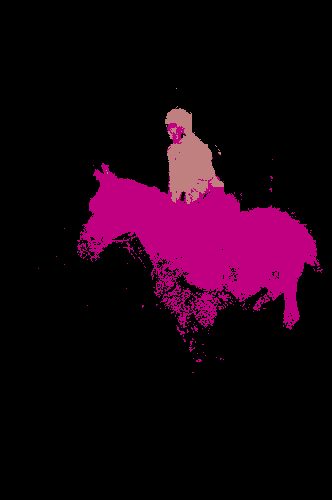} &
\includegraphics[width=.09\textwidth]{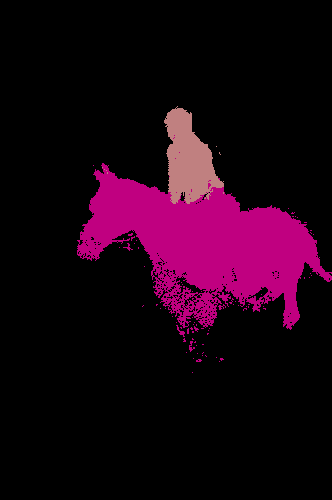} &
\includegraphics[width=.09\textwidth]{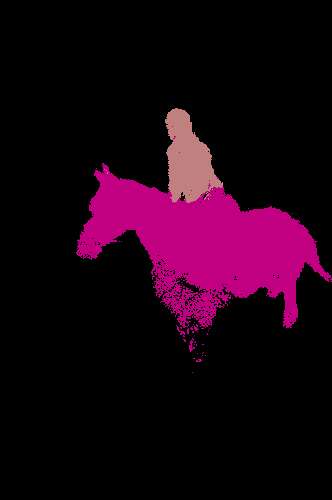} & \includegraphics[width=.09\textwidth]{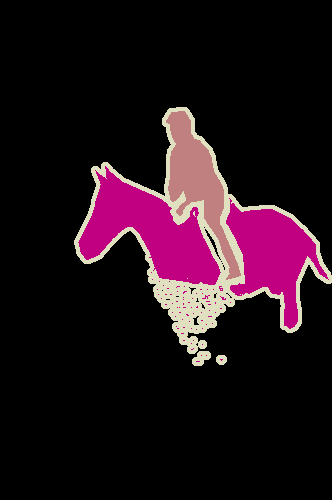}\\

\includegraphics[width=.09\textwidth]{Results/2011_001232_img.png} &
\includegraphics[width=.09\textwidth]{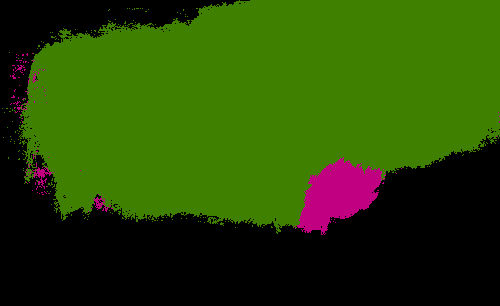} &
\includegraphics[width=.09\textwidth]{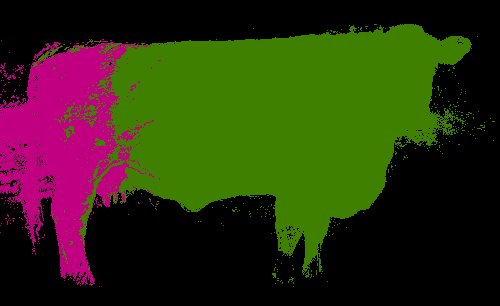} &
\includegraphics[width=.09\textwidth]{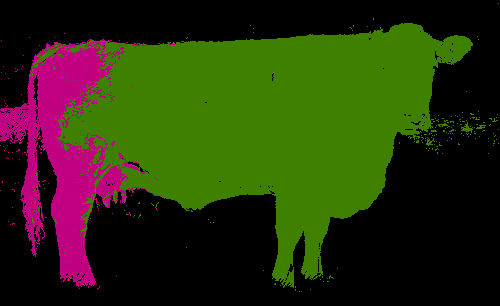} &
\includegraphics[width=.09\textwidth]{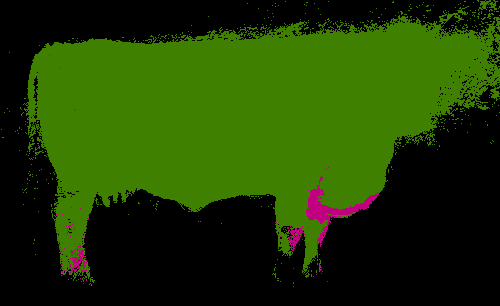} &
\includegraphics[width=.09\textwidth]{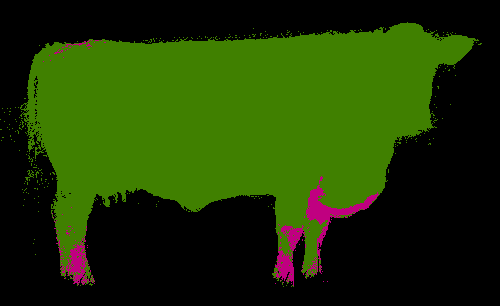} &
\includegraphics[width=.09\textwidth]{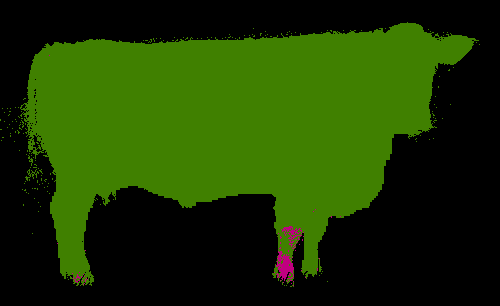} & \includegraphics[width=.09\textwidth]{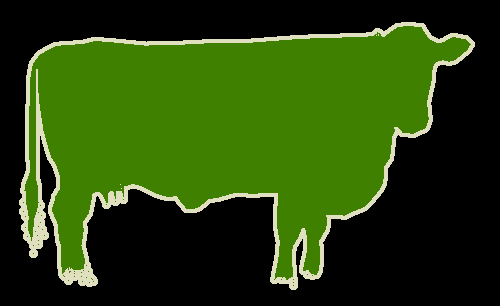}\\

\end{tabular}
\caption{Qualitative results from the Pascal VOC validation set.}
\label{fig:success_result}
\end{figure*}

\subsection{Comparison to State-of-the-art Methods}
\label{sec:comp}
We first compare our approach with state-of-the-art baselines on PASCAL VOC. To this end, we report the Intersection over Union (IOU), which is the most commonly used metric for semantic segmentation.
In the following, we refer to our approach with foreground/background masks as \emph{Ours fg/bg masks} and with multi-class masks as \emph{Ours multi-class masks}.

We report the results of our approach and the state-of-the-art methods relying on tags only in Table~\ref{tab:pascal_tags} and Table~\ref{tab:pascal_test} for the Pascal VOC 2012 validation and test images, respectively.
Note that our approach, with either type of masks, outperforms most of the baselines by a large margin. The only exception is the contemporary SEC algorithm of~\cite{SEC}, which outperforms the foreground/background version of our method. Note that SEC also relies on the multi-class results of the localization network. We believe that the fact that our multi-class-based approach performs slightly better than SEC, particularly on the test images, indicates the effectiveness of our combination of the localization network with our fusion-based built-in prior. Importantly, the results also show that we outperform the methods based on an objectness prior~\cite{whatsthepoint,fromimgtopixel}, which evidences the benefits of using our built-in foreground/background masks instead of external objectness algorithms. Note that our results with foreground/background masks vary slightly from those reported in our previous paper~\cite{OurECCV}, due to changes in the CRF parameters and the use of higher-order potentials.

\begin{figure*}[!t]
\centering
\tiny
\begin{tabular}{cccccccc}
 Image & baseline & fg/bg mask & fg/bg mask+HO & multi-class mask/Loc & multi-class mask/Loc &multi-class mask/Loc& GT  \\
  &  &  &  & & +Fusion &+Fusion+HO &   \\
  
\includegraphics[width=.09\textwidth]{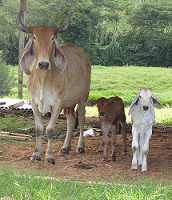} &
\includegraphics[width=.09\textwidth]{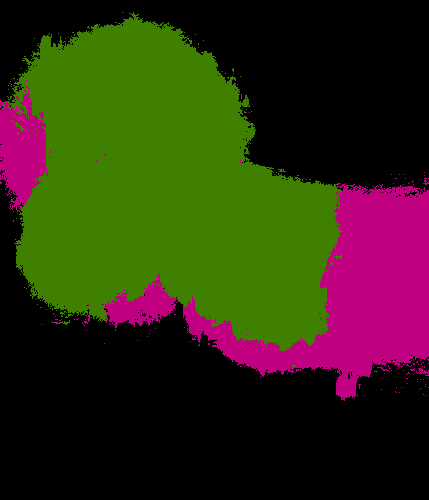} &
\includegraphics[width=.09\textwidth]{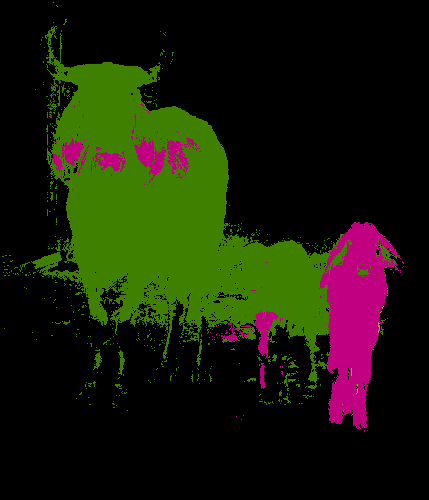} &
\includegraphics[width=.09\textwidth]{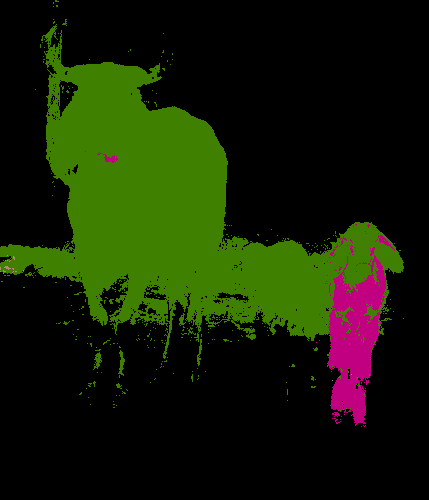} &
\includegraphics[width=.09\textwidth]{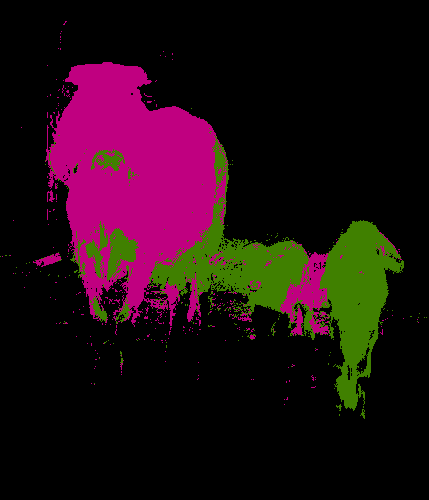} &
\includegraphics[width=.09\textwidth]{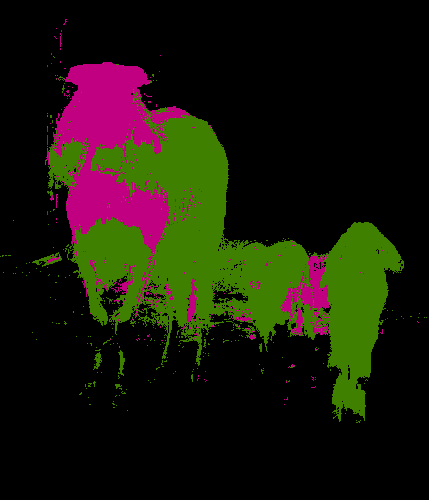} &
\includegraphics[width=.09\textwidth]{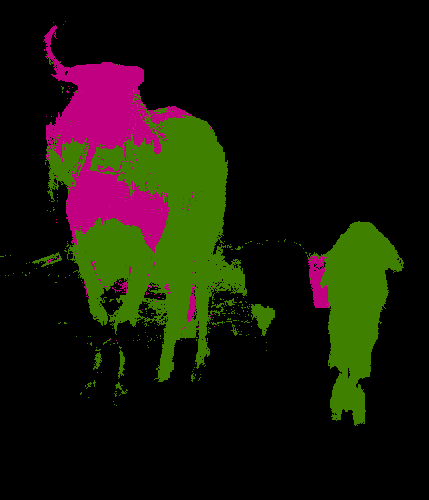} & \includegraphics[width=.09\textwidth]{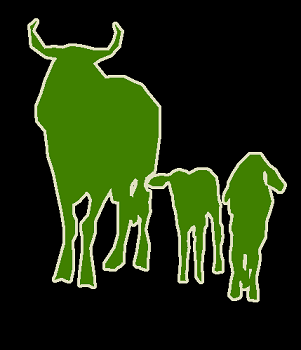}\\

\includegraphics[width=.09\textwidth]{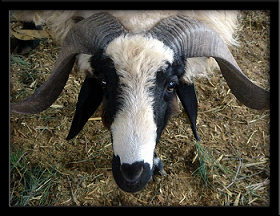} &
\includegraphics[width=.09\textwidth]{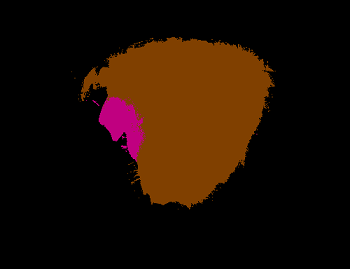} &
\includegraphics[width=.09\textwidth]{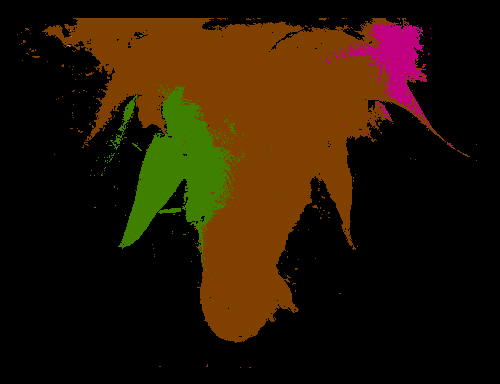} &
\includegraphics[width=.09\textwidth]{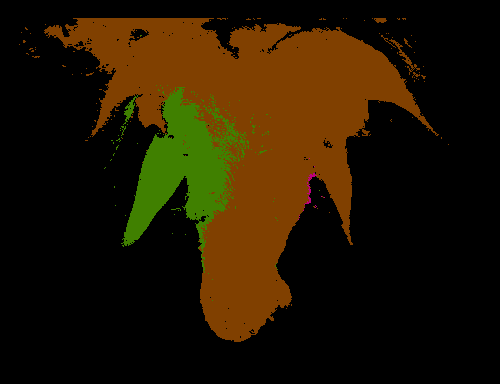} &
\includegraphics[width=.09\textwidth]{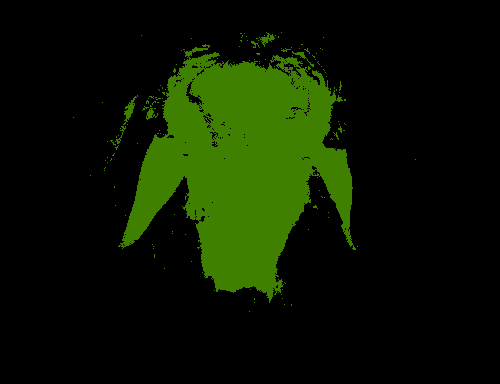} &
\includegraphics[width=.09\textwidth]{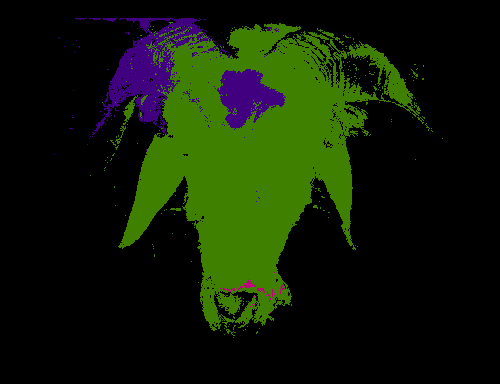} &
\includegraphics[width=.09\textwidth]{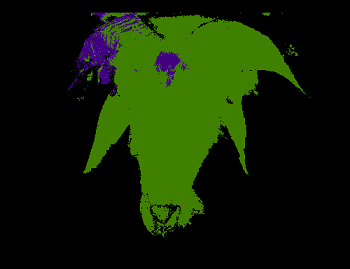} & \includegraphics[width=.09\textwidth]{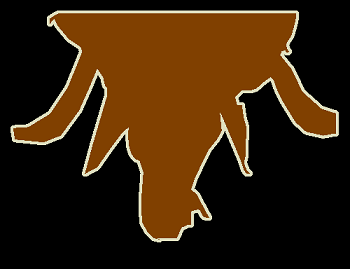}\\

\includegraphics[width=.09\textwidth]{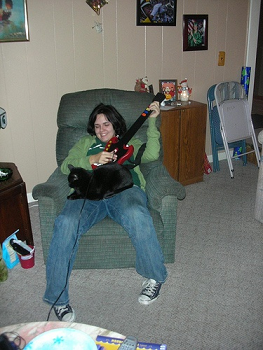} &
\includegraphics[width=.09\textwidth]{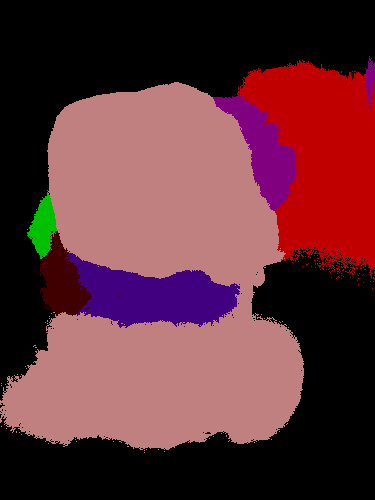} &
\includegraphics[width=.09\textwidth]{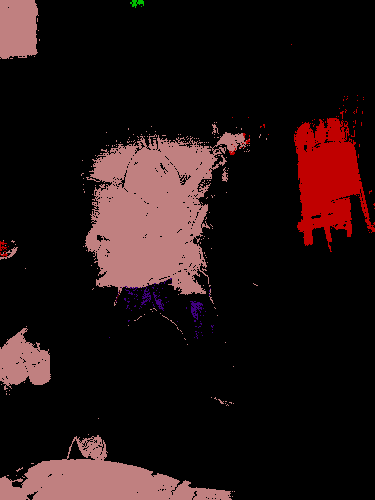} &
\includegraphics[width=.09\textwidth]{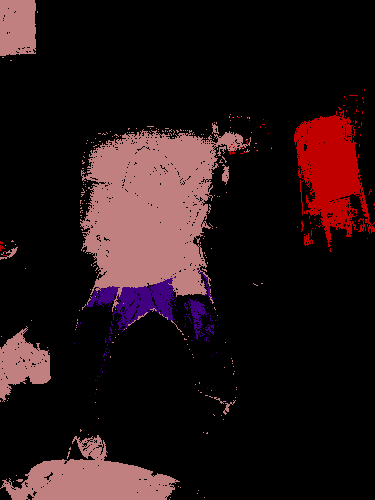} &
\includegraphics[width=.09\textwidth]{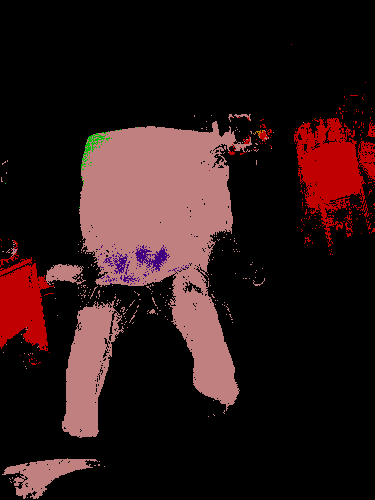} &
\includegraphics[width=.09\textwidth]{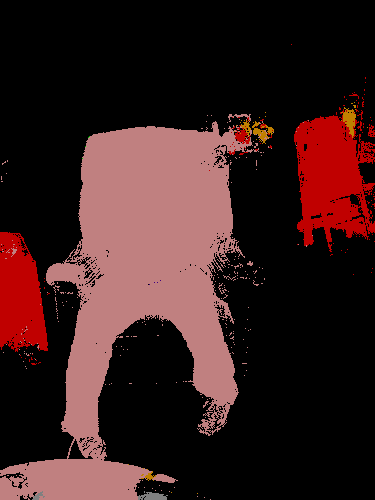} &
\includegraphics[width=.09\textwidth]{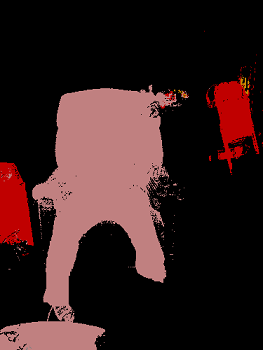} & \includegraphics[width=.09\textwidth]{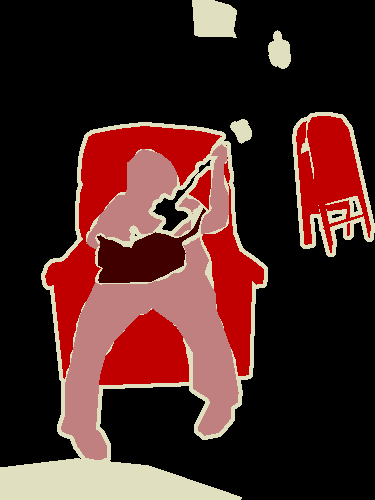}\\

\includegraphics[width=.09\textwidth]{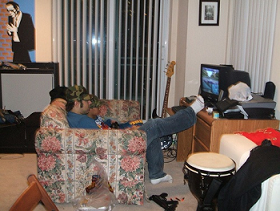} &
\includegraphics[width=.09\textwidth]{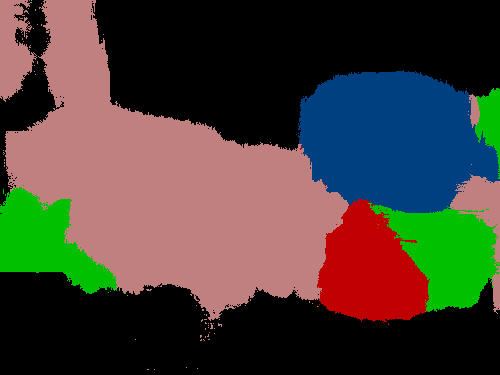} &
\includegraphics[width=.09\textwidth]{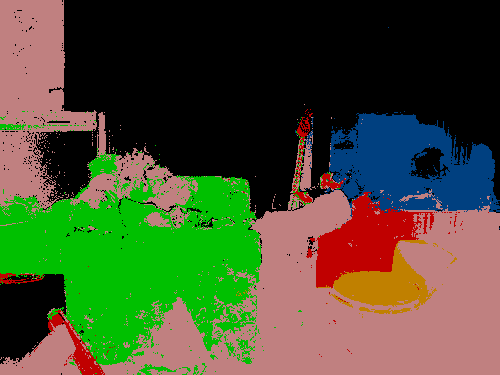} &
\includegraphics[width=.09\textwidth]{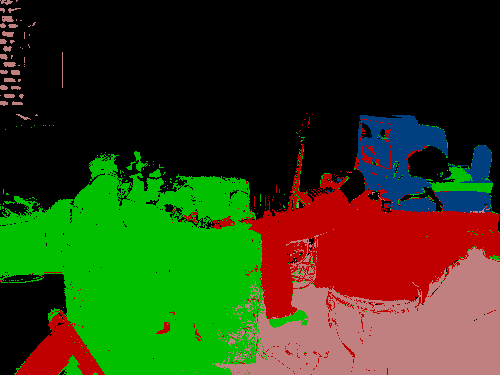} &
\includegraphics[width=.09\textwidth]{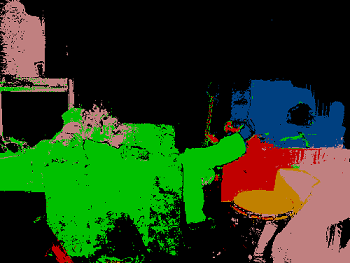} &
\includegraphics[width=.09\textwidth]{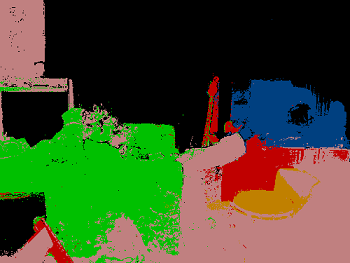} &
\includegraphics[width=.09\textwidth]{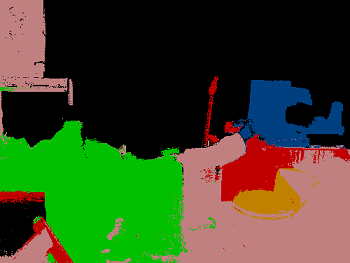} & \includegraphics[width=.09\textwidth]{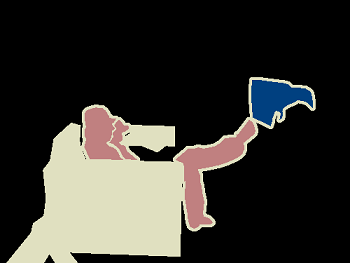}\\
  \end{tabular}
\label{fig:failure_result}
\end{figure*}

\begin{figure*}[!t]
\centering
\tiny
\begin{tabular}{cccccccccccccccccccc}

\includegraphics[width=.02\textwidth]{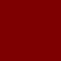} &
\includegraphics[width=.02\textwidth]{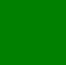} &
\includegraphics[width=.02\textwidth]{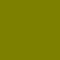} &
\includegraphics[width=.02\textwidth]{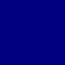} &
\includegraphics[width=.02\textwidth]{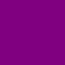} &
\includegraphics[width=.02\textwidth]{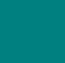} &
\includegraphics[width=.02\textwidth]{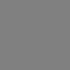} &
\includegraphics[width=.02\textwidth]{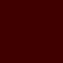}&
\includegraphics[width=.02\textwidth]{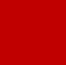}&
\includegraphics[width=.02\textwidth]{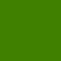}&
\includegraphics[width=.02\textwidth]{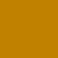}&
\includegraphics[width=.02\textwidth]{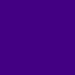}&
\includegraphics[width=.02\textwidth]{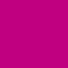}&
\includegraphics[width=.02\textwidth]{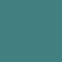}&
\includegraphics[width=.02\textwidth]{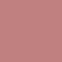}&
\includegraphics[width=.02\textwidth]{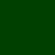}&
\includegraphics[width=.02\textwidth]{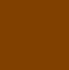}&
\includegraphics[width=.02\textwidth]{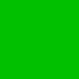}&
\includegraphics[width=.02\textwidth]{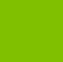}&
\includegraphics[width=.02\textwidth]{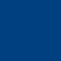}\\
 \rotatebox[origin=c]{90}{aero} & \rotatebox[origin=c]{90}{bike} & \rotatebox[origin=c]{90}{bird} & \rotatebox[origin=c]{90}{boat} & \rotatebox[origin=c]{90}{bottle} & \rotatebox[origin=c]{90}{bus} & \rotatebox[origin=c]{90}{car} & \rotatebox[origin=c]{90}{cat} & \rotatebox[origin=c]{90}{chair} & \rotatebox[origin=c]{90}{cow} & \rotatebox[origin=c]{90}{table} & \rotatebox[origin=c]{90}{dog} & \rotatebox[origin=c]{90}{horse} & \rotatebox[origin=c]{90}{mbike} & \rotatebox[origin=c]{90}{person} & \rotatebox[origin=c]{90}{plant} & \rotatebox[origin=c]{90}{sheep}& \rotatebox[origin=c]{90}{sofa} & \rotatebox[origin=c]{90}{train} & \rotatebox[origin=c]{90}{tv}  \\
  \end{tabular}
  \caption{Failure cases from the Pascal VOC validation set.}
\label{fig:failure_result}
\end{figure*}

We then compare our approach, which uses only image tags, with other methods that rely on additional training data or additional supervision. In particular, these include the point supervision of~\cite{whatsthepoint}, the random crops of~\cite{weaklyandsemi}, the size information of~\cite{CCNN}, the MCG segments of~\cite{fromimgtopixel,learnTosegwithimagelevelanno,augment_feedback}, additional training data of~\cite{STC}, and the CheckMask procedure of our previous work~\cite{OurECCV}. The results of this comparison are provided in Table~\ref{tab:pascal_sup}. Note that, with the exception of our own CheckMask procedure and the method of~\cite{augment_feedback}, which uses MCG segments, our approach with multi-class masks outperforms all the baselines, and with foreground/background masks most of them, despite the fact that we do not require any supervision other than tags. It is worth mentioning that other approaches have proposed to rely on labeled bounding boxes, which require a user to provide a bounding box for each individual foreground object in an image and to associate a label to each such bounding box. While this procedure is clearly costly, we achieve accuracies close to these baselines (52.5\% for~\cite{weaklyandsemi} when using labeled bounding boxes and 54.1\% for~\cite{weaklyandsemi} when using labeled bounding boxes in an EM process vs 50.9\% for our approach with image tags only). We believe that this further evidences the benefits of our approach.

In Figs.~\ref{fig:success_result} and~\ref{fig:failure_result}, we show some successful segmentations and failure cases of our approach, respectively. In some cases (e.g., first row of Fig.~\ref{fig:failure_result}), these failures are due to the output scores of the network, which are used in Eqs.~\ref{eq:loss_mil_mask} and~\ref{eq:loss_mil_multimask}. Other failures are due to errors in our predicted masks. For example, the second row of Fig.~\ref{fig:failure_result} indicates that errors appear after using the localization network to generate multi-class masks. The most common type of failure occurs in the presence of complex scenes, in which the network is unable to segment small objects. The last two rows of Fig.~\ref{fig:failure_result} show some of these cases. 

\subsection{Ablation Study}
\label{sec:ablation}
We now study the effect of the different components of our approach on our results. In particular, we first evaluate our predicted masks, and then discuss semantic segmentation results.

\subsubsection{Mask Evaluation}

{\bf Foreground/background masks.}
To evaluate our foreground/background masks, we made use of 10\% of randomly chosen training images from the Pascal VOC dataset. We then generated foreground/background masks for these images using our approach, which relies on the activations of the fourth and fifth layers of the segmentation network pre-trained on ImageNet (i.e., before fine-tuning it for semantic segmentation). These masks can then be compared to ground-truth foreground/background masks obtained directly from the pixel level annotations.

We compare our masks with the objectness criteria of~\cite{objectness} and~\cite{MCG}, which were employed for the purpose of weakly-supervised semantic segmentation by~\cite{whatsthepoint} and~\cite{fromimgtopixel,learnTosegwithimagelevelanno,augment_feedback}, respectively. Note that some objectness methods, such as~\cite{BING,MCG}, that have been used for weakly-supervised semantic segmentation~\cite{fromimgtopixel,boxsup,learnTosegwithimagelevelanno,augment_feedback}, require training data with pixel-level or bounding box annotations, and thus are not really comparable to our approach. Note also that a complete evaluation of objectness methods goes beyond the scope of this paper, which focuses on weakly-supervised semantic segmentation. 

\begin{table}[t!]
\renewcommand{\arraystretch}{1.2}
\centering
\caption{Comparison of our foreground/background masks with those obtained using the objectness methods of~\cite{objectness} and~\cite{MCG}.}
\label{tab:objectness}
\begin{tabular}{l|c}
 & Mean IoU\\
\hline
Masks obtained using~\cite{objectness} & 52.34\% \\
Masks obtained using~\cite{MCG} & 50.20\% \\
Our masks & {\bf 60.08}\% \\
\end{tabular}
\end{table}

\begin{figure}[!t]
\centering
\tiny
\begin{tabular}{c c c c}
Image & Ground-Truth & Localization map & Average on GT\\
\hline
Success for monitor\\
\hline
\includegraphics[width=.1\textwidth]{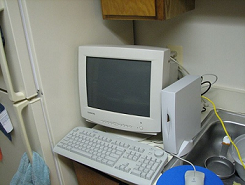} &
\includegraphics[width=.1\textwidth]{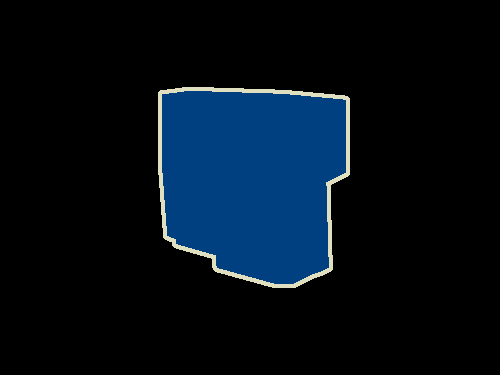} & 
\includegraphics[width=.1\textwidth]{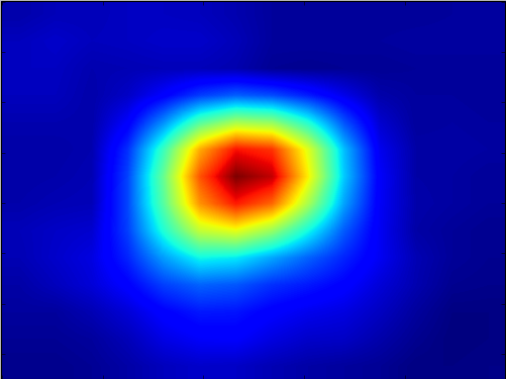} \\
\includegraphics[width=.1\textwidth]{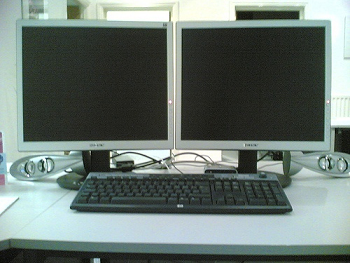} &
\includegraphics[width=.1\textwidth]{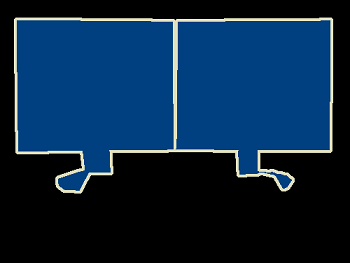} &
\includegraphics[width=.1\textwidth]{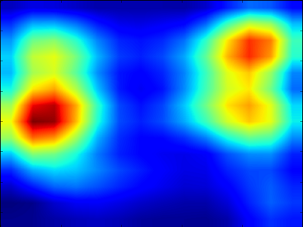} &
\includegraphics[width=.08\textwidth]{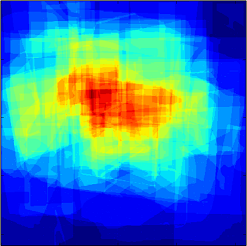} \\
\includegraphics[width=.1\textwidth]{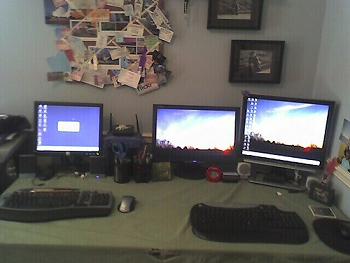} &
\includegraphics[width=.1\textwidth]{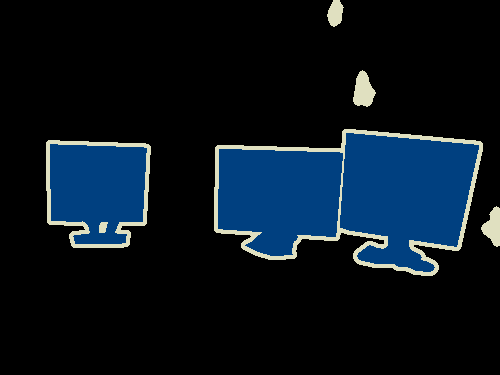} &
\includegraphics[width=.1\textwidth]{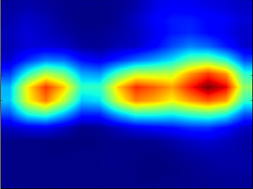} \\
\hline
Failure for potted plant\\
\hline
\includegraphics[width=.1\textwidth]{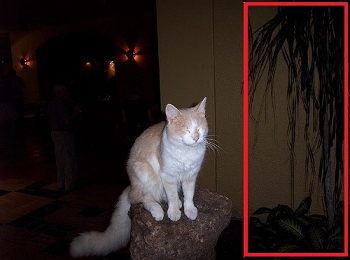} &
\includegraphics[width=.1\textwidth]{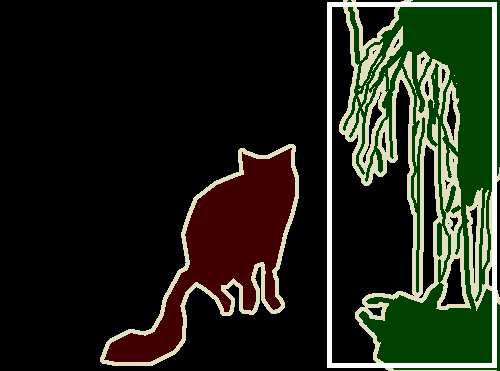} & 
\includegraphics[width=.1\textwidth]{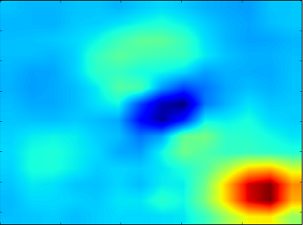} \\
\includegraphics[width=.1\textwidth]{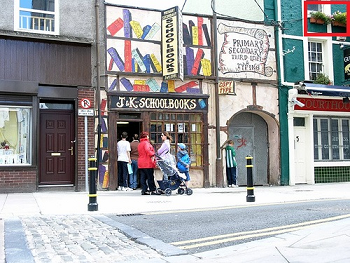} &
\includegraphics[width=.1\textwidth]{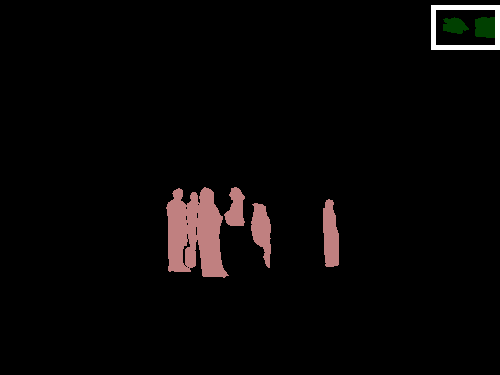} &
\includegraphics[width=.1\textwidth]{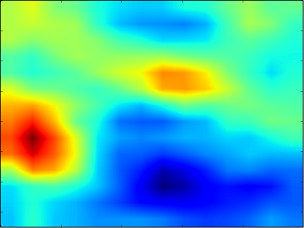} &
\includegraphics[width=.08\textwidth]{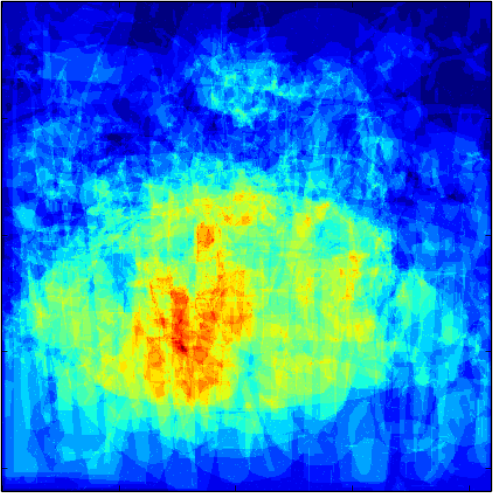}\\
\includegraphics[width=.1\textwidth]{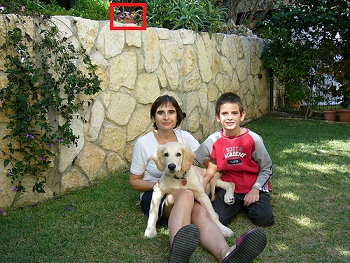} &
\includegraphics[width=.1\textwidth]{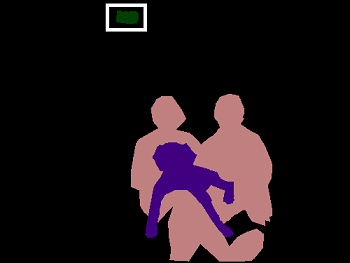} &
\includegraphics[width=.1\textwidth]{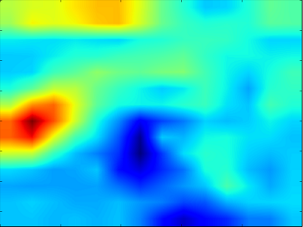} \\
\end{tabular}
\caption{Success and failure cases of the localization network.} 
\label{fig:only_localization}
\end{figure}

The objectness methods of~\cite{objectness} and~\cite{MCG} produce a per-pixel foreground probability map. For our comparison to be fair, we further refined these maps using the same dense CRF as in our approach. In Table~\ref{tab:objectness}, we provide the results of these experiments in terms of mean Intersection Over Union (mIOU) with respect to the ground-truth masks. Note that our masks are more accurate than those of~\cite{objectness,MCG}. 
In particular, we have found that our masks yield a much better object localization accuracy.

{\bf Multi-class masks.}
As discussed before, our multi-class masks rely, in part, on the localization network. Although the localization map provides useful information about the location of the objects, it is not sufficient on its own to generate accurate masks. In addition to its lack of accuracy at the object's boundary and the incompleteness of its segmentation, as illustrated earlier in Fig.~\ref{fig:loc_only}, the accuracy of the localization network varies greatly for different classes. We illustrate this in Fig.~\ref{fig:only_localization} for the successful case of the monitor class and for the failure case of potted plants. In the case of monitor, which, most of the time, is located in the center of the image (see Average on GT), the network is able to localize it reasonably well. By contrast, potted plants are scattered in all locations in the dataset (see Average on GT), and the network therefore fails to localize it accordingly. As a matter of fact, when training our method with masks obtained from the localization network only, the IOU of potted plants is 0. This IOU increases to 32.3 when combining the localization network with our fusion-based masks, as discussed in Section~\ref{sec:mc}.

\begin{table}[!t]
\renewcommand{\arraystretch}{1.2}
\centering
\caption{Accuracy of the multi-class masks when directly used for segmentation (without any network), assuming known tags at test time.}
\label{tab:mask_eval}
\begin{tabular}{l|c}
 Methods & Mean IoU \\
\hline
multi-class masks using localization & 43.0 \\
multi-class masks using localization+fusion & 46.6 \\
\end{tabular}
\end{table}

\begin{figure}[!t]
\centering
\includegraphics[width=0.5\textwidth]{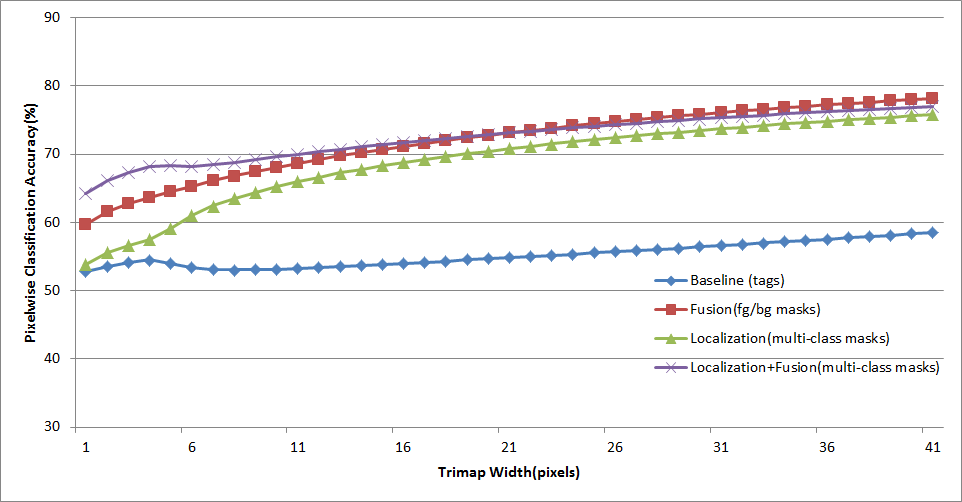}
\caption{Pixel classification accuracy as a function of the bandwidth around the object boundaries on the Pascal VOC validation set. Note that using our fusion-based masks helps improving the accuracy at the boundary of the objects.}
\label{fig:trimap}
\end{figure}

Since our method generates multi-class masks, one could think of directly using these masks to obtain the final semantic segmentation of an input image, that is, without training a network at all. We evaluated how well this naive approach performs on the Pascal VOC validation data. To further help this baseline, we made use of the ground-truth tags to filter out the absent classes from the masks' predictions. The results of this experiment are reported in Table~\ref{tab:mask_eval} for the localization masks only and for our multi-class masks. Note that these results, despite relying on ground-truth tags at test time, are lower than that of our approach, which does not use this information. This confirms the importance of training a network based on our masks, rather than directly using the masks for prediction.

To evaluate the accuracy of the different types of masks at the boundary of the objects, we further made use of the Trimap accuracy~\cite{trimap}, which focuses on the segmentation error within a region around the true boundaries. In Fig.~\ref{fig:trimap}, we report the Trimap accuracy as a function of the width of the region around the boundary for the results obtained with our fusion-based foreground/background masks, the localization network masks, and our multi-class masks (fusion+localization). In addition to this, we also report the error of a simple baseline consisting of not using any mask, but only the tags, i.e., using Eq.~\ref{eq:loss_mil} as training loss. Note that using masks clearly improves boundary accuracy, particularly when using our fusion-based masks, with or without the additional localization ones. Recall, however, that the combination of fusion+localization gave higher accuracy than fusion only in terms of IOU. This shows the benefits of our complete multi-class masks.

\begin{table}
\renewcommand{\arraystretch}{1.2}
\centering
\caption{Mean IOU on PASCAL VOC val. set for different setups of our method.}
\label{tab:pascal_setup}
\begin{tabular}{l|c}
 Methods & mIOU\\
\hline
Tag-only Baseline (no mask) & 31.0 \\
Foreground/Background Priors & 47.3 \\
Foreground/Background Priors + Higher Order & 48.0 \\
Localization Priors & 45.9 \\
Localization Priors+Higher Order & 46.6\\
Localization+Fusion Priors & 49.2\\
Localization+Fusion Priors+Higher Order (small FOV)& 49.3 \\
Localization+Fusion Priors+Higher Order (large FOV) & 50.9 \\
\end{tabular}
\end{table}

\subsubsection{Effect of the Different Components}

In Table~\ref{tab:pascal_setup}, we evaluate the influence of several components of our approach. In particular, we report the results of the simple baseline mentioned above that only uses tags, but no mask. 
We also report the results obtained with different types of masks, with and without using the higher-order terms in our CRF smoothing procedure, and, in the multi-class case, with different network fields-of-view. 
The importance of our mask is clearly evidenced by the fact that mask-based results outperform the mask-free baseline by up to 17.0 mIOU points when using foreground/background masks and up to 19.9 when using multi-class masks. These results also show that using higher-order terms brings some improvement over the pairwise CRF, albeit of much lesser magnitude than the masks themselves. Similarly, the network field-of-view has some influence on accuracy. 

\textbf{Computation time}. For each validation image of PASCAL VOC, the network forward time is of 0.06 sec. on an NVIDIA TESLA P100 GPU.
The running time of crisp boundary detection for a single image takes 4.1 seconds when using the speedy version of the public Matlab implementation~\cite{Crisp} on a single core of an Intel Core i5 processor. For the Dense CRF, inference takes 2.8 and 2.1 seconds with and without higher-order term, respectively, using the public C++ code~\cite{vineet_higherorder} on a single core of an Intel Core i7 processor. The bottleneck of our approach at test time therefore is the crisp boundary detection. Note, however, that this step is only used to determine the regions for the higher-order potentials, without which, as shown in Table~\ref{tab:pascal_setup}, our approach still yields competitive results.

\subsection{Evaluation on YTO and MS COCO}
\label{sec:coco}
To further demonstrate the generality of our method, we conducted a set of experiments on YTO and MS COCO. While a few weakly-supervised methods have been applied to YTO, to the best of our knowledge, no weakly supervised results have been published on MS-COCO. We therefore also computed the results of the contemporary SEC method~\cite{SEC} on these two datasets using the publicly available code.

\subsubsection{Evaluation on YTO}
In Table~\ref{tab:yto_res} we report the per class mean IOU of our approach and several baselines on YTO.
Note that our method outperforms all the baselines, including~\cite{SEC}, on this dataset.

\begin{table}[!t]
\renewcommand{\arraystretch}{1.2}
\centering
\caption{Per class IOU on Youtube Objects using image tags during training.}
\label{tab:yto_res}
\scriptsize
\scalebox{0.685}
{
\begin{tabular}{| l|c c c c c c c c c c c |c |} 
\hline
Method & \rotatebox[origin=c]{90}{bg} & \rotatebox[origin=c]{90}{aero} & \rotatebox[origin=c]{90}{bird} & \rotatebox[origin=c]{90}{boat} & \rotatebox[origin=c]{90}{car} & \rotatebox[origin=c]{90}{cat} & \rotatebox[origin=c]{90}{cow} & \rotatebox[origin=c]{90}{dog} & \rotatebox[origin=c]{90}{horse} & \rotatebox[origin=c]{90}{mbike} & \rotatebox[origin=c]{90}{train} & \rotatebox[origin=c]{0}{mIOU}\\
 \hline
Papazoglou et al.~\cite{fast} &-& 67.4 & 62.5 & 37.8  & 67.0 & 43.5 & 32.7 & 48.9 & 31.3 & 33.1 & 43.4& 46.8\\
Tang et al.~\cite{discriminative} &-& 17.8 & 19.8 & 22.5 & 38.3 & 23.6 & 26.8 & 23.7 & 14.0& 12.5 & 40.4 & 23.9\\
Ochs et al.~\cite{long-term-video}&-&13.7 & 12.2 & 10.8 & 23.7 & 18.6 & 16.3 & 18.0 & 11.5 & 10.6&  19.6 & 15.5\\
SEC~\cite{SEC} & 84.4 & 51.9 & 59.3 & 37.5 & 64.4 & 30.5 & 38.2 & 50.1 & 51.1 & 49.7 & 17.3 &  48.6\\
Ours (multi-class masks) & 88.5 & 72.7 & 60.1 & 44.2 & 53.5 & 33.3 & 42.4 & 50.3 & 49.6 & 56.6 & 16.6 &  51.6\\
\hline
\end{tabular}
}
\end{table}

\subsubsection{Evaluation on MS COCO}
MS COCO is a large-scale dataset containing 80 classes of different categories. Unlike in PASCAL VOC and YTO, in MS-COCO, the majority of samples were collected from non-iconic images in a complex natural context. Moreover, a large number of the classes, e.g., spoon and knife, are \emph{small} in terms of both size and the number of instances/samples in the datasets. Additionally, the classes of similar categories, e.g., Furniture and Indoor categories, appear together in an image, resulting in images depicting more than 10 classes. 
These properties make MS-COCO very challenging for weakly-supervised segmentation, and, to the best of our knowledge, we are the first to report results on this dataset in the weakly-supervised setting.

\begin{table}[!t]
\renewcommand{\arraystretch}{1.1}
\centering
\scriptsize
\caption{Per class IOU on MS COCO using image tags during training.}
\label{tab:coco}
\scalebox{0.9}
{
\hspace{-0.2cm}\begin{tabular}{|c| l|c| c || c|l| c |c|} 
\hline
Cat. & Class & \rotatebox[origin=c]{0}{SEC} & \rotatebox[origin=c]{0}{Ours} & Cat. & \rotatebox[origin=c]{0}{Class} & \rotatebox[origin=c]{0}{SEC} & \rotatebox[origin=c]{0}{Ours} \\
 \hline
BG& background & 74.3 & 68.8 & \multirow{6}{*}{\rotatebox[origin=c]{90}{Kitchenware}}& wine glass & 22.3&17.5 \\
\cline{1-4}
P&person &43.6 & 27.5& &cup & 17.9&5.6\\
\cline{1-4}
\multirow{8}{*}{\rotatebox[origin=c]{90}{Vehicle}}& bicycle & 24.2& 18.2 & & fork & 1.8&0.5 \\
& car & 15.9& 7.2& & knife&1.4&1.0\\
& motorcycle & 52.1& 40.5& & spoon&0.6&0.6\\
& airplane & 36.6& 32.0& & bowl& 12.5&13.3\\\cline{5-8}
& bus & 37.7& 39.2& \multirow{10}{*}{\rotatebox[origin=c]{90}{Food}} & banana& 43.6&44.9\\
& train & 30.1& 26.5& & apple& 23.6&18.9\\
& truck & 24.1& 17.5& & sandwich& 22.8&21.4\\
& boat & 17.3& 16.5& & orange& 44.3&35.0\\
\cline{1-4}
\multirow{6}{*}{\rotatebox[origin=c]{90}{Outdoor}}& traffic light & 16.7& 3.9& & broccoli& 36.8&27.0\\
&fire hydrant & 55.9& 33.1& & carrot& 6.7&16.0\\
&stop sign & 48.4& 28.4& & hot dog& 31.2&22.5\\
&parking meter & 25.2& 25.5& & pizza& 50.9&57.8\\
&bench & 16.4& 12.4& & donut& 32.8&36.2\\
\cline{1-4}
\multirow{10}{*}{\rotatebox[origin=c]{90}{Animal}}& bird & 34.7& 31.1& & cake& 12.0&17.0\\\cline{5-8}
& cat & 57.2& 52.8&\multirow{6}{*}{\rotatebox[origin=c]{90}{Furniture}} & chair& 7.8&8.2\\
& dog & 45.2& 44.1& & couch& 5.6&13.9\\
& horse & 34.4& 34.2& & potted plant& 6.2&7.4\\
& sheep & 40.3& 38.0& & bed& 23.4&29.8\\
& cow & 41.4& 42.1& & dining table& 0.0&2.0\\
& elephant & 62.9& 65.2& & toilet& 38.5&30.1\\\cline{5-8}
& bear & 59.1& 57.0& \multirow{6}{*}{\rotatebox[origin=c]{90}{Electronics}} & tv& 19.2&14.8\\
& zebra & 59.8& 65.0& & laptop& 20.1&19.9\\
& giraffe & 48.8& 55.6& & mouse& 3.5&0.4\\
\cline{1-4}
\multirow{5}{*}{\rotatebox[origin=c]{90}{Accessory}}& backpack & 0.3& 3.2& & remote& 17.5&9.9\\
& umbrella & 26.0& 28.1& & keyboard& 12.5&19.9\\
& handbag & 0.5& 1.1& & cell phone& 32.1&26.1\\\cline{5-8}
& tie & 6.5& 5.5& \multirow{5}{*}{\rotatebox[origin=c]{90}{Appliance}} & microwave& 8.2&9.8\\
& suitcase & 16.7& 21.3& & oven& 13.7&16.4\\
\cline{1-4}
\multirow{10}{*}{\rotatebox[origin=c]{90}{Sport}}& frisbee & 12.3& 5.6& & toaster& 0.0&0.0\\
& skis & 1.6& 1.0& & sink& 10.8&9.5\\
& snowboard & 5.3& 2.8& & refrigerator& 4.0&13.2\\\cline{5-8}
& sports ball & 7.9& 1.9& \multirow{7}{*}{\rotatebox[origin=c]{90}{Indoor}} & book& 0.4&7.5\\
& kite & 9.1& 10.3& & clock& 17.8&16.5\\
& baseball bat &1.0& 1.7& & vase& 18.4&13.4\\
& baseball glove & 0.6& 0.5& & scissors& 16.5&12.2\\
& skateboard & 7.1& 6.6& & teddy bear& 47.0&41.0\\
& surfboard & 7.7& 3.3& & hair dryer& 0.0&0.0\\
& tennis racket & 9.1& 5.5& & toothbrush& 2.8&2.0\\\cline{5-8}
\cline{1-4}
& bottle & 13.2& 9.6& & \textbf{mean IOU}&22.4 & 20.4 \\
\hline
\end{tabular}
}
\end{table}

\begin{figure}
\centering
\begin{tabular}{c}
\includegraphics[width=0.4\textwidth]{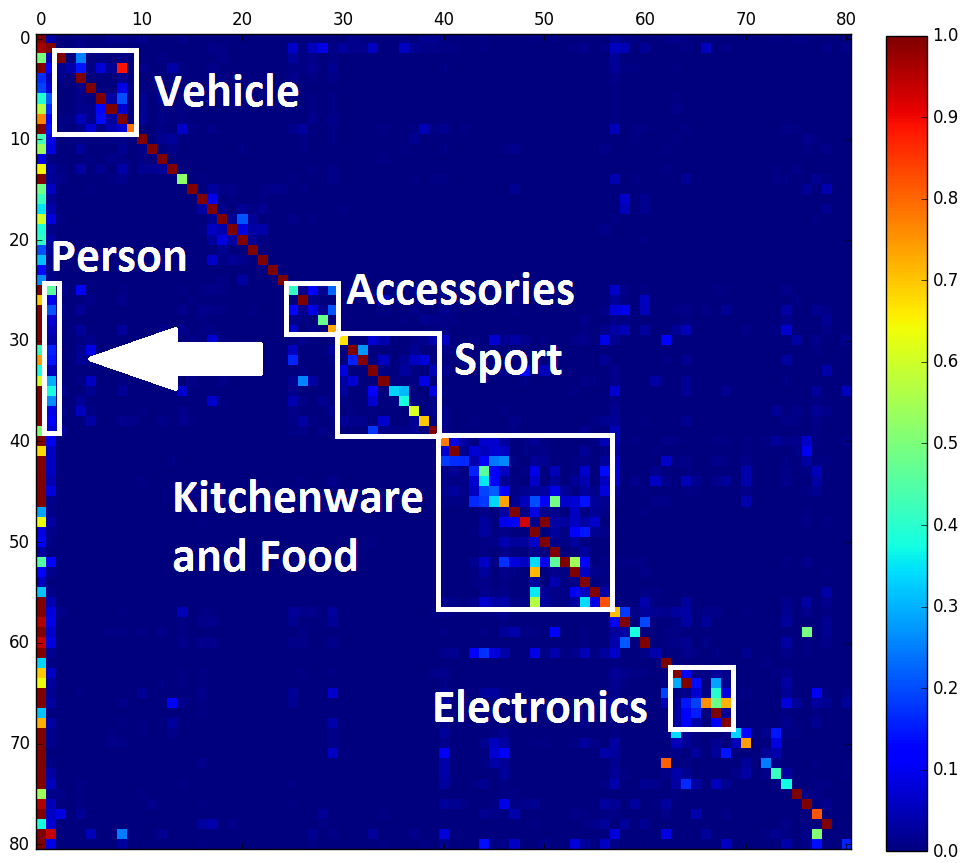}
 \\
\end{tabular}

\caption{Confusion matrix of our method on the MS COCO validation set. The classes are shown in the same order as in Table~\ref{tab:coco}. Note that the main sources of confusion are with the background or with classes coming from the same broad category or appearing in the same context.}
\label{fig:coco_conf}
\end{figure}

We provide the per-class IoU of our approach and SEC~\cite{SEC} in Table~\ref{tab:coco}. While, on average, SEC obtains slightly better results, the behavior of both methods is similar: They yield reasonable accuracy on large classes, such as Animals, but perform poorly on small ones, such as Indoor and Kitchenware. Interestingly, by analyzing the confusion matrix depicted in Fig.~\ref{fig:coco_conf}, we noticed that our approach is more confused between classes from the same broad category. For instance, there are large confusions between the classes of the 'Food and Kitchenware' category.
Furthermore, many of the classes from accessories and sport is confused with Person as in most samples they appear together with Person.

Altogether, we believe that, while promising, these results on MS-COCO evidence that there is much space for progress in weakly-supervised semantic segmentation, and, in particular, that developing solutions that improve intra-category discrimination could be an interesting direction for future research.

\section{Conclusion}
We have introduced a Deep Learning approach to weakly-supervised semantic
segmentation that leverages masks directly extracted
from networks pre-trained for the task of object recognition. In particular, we have shown how to extract foreground/background masks by fusing the activations of convolutional layers, as well as multi-class ones by combining this fusion-based prior with a localization one. Our experiments have shown the benefits of our masks, and in particular of the multi-class ones, which yield state-of-the-art segmentation accuracy on PASCAL VOC. The most common failure cases of our approach are related to the presence of small objects in complex scenes. In the future, we will therefore focus on addressing this issue. Furthermore, a general limitation of existing tag-based semantic segmentation techniques is that they assume that the tags cover all the classes in the input image. We believe that developing algorithms that go beyond this fairly unrealistic assumption is a promising research direction.



\ifCLASSOPTIONcaptionsoff
  \newpage
\fi

\bibliographystyle{IEEEtran}
\bibliography{egbib}

\begin{IEEEbiography}[{\includegraphics[width=1in,height=1.25in,clip,keepaspectratio]{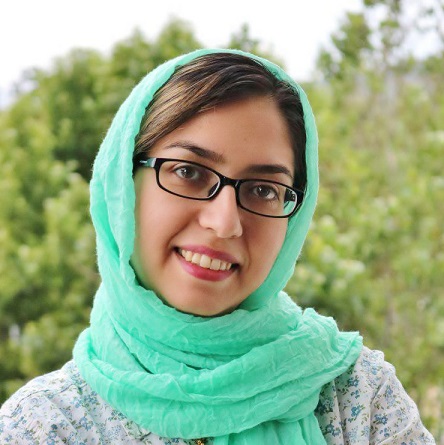}}]{Fatemehsadat Saleh}
obtained her BSc from Isfahan University of Technology, Iran and MSc from Sharif University of Technology, Iran in 2011 and 2013, respectively. She then joined Electronic Research Institute at Sharif University of Technology as a researcher. She is now a PhD student at the Australian National University and Data61 of CSIRO at Canberra research lab. Her main research interests are in scene understanding of images and videos, especially in a weakly-supervised manner.
\end{IEEEbiography}

\vspace{-20pt}
\begin{IEEEbiography}[{\includegraphics[width=1in,height=1.25in,clip,keepaspectratio]{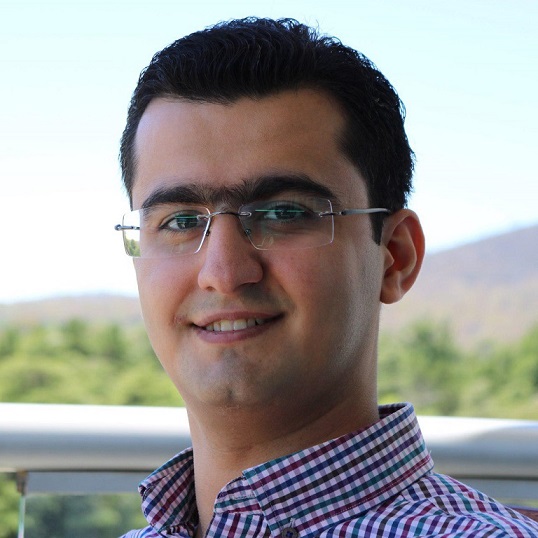}}]{Mohammad Sadegh Aliakbarian}
obtained his BSc degree from Isfahan University of Technology in 2013. He was with Electronic Research Institute at Sharif University of Technology, Tehran, Iran as a researcher and developer and then joined NICTA in Canberra as a visitor researcher. He is now a PhD student at the Australian National University and Data61 of CSIRO at Canberra research lab. 
\end{IEEEbiography}

\vspace{-20pt}
\begin{IEEEbiography}[{\includegraphics[width=1in,height=1.2in,clip,keepaspectratio]{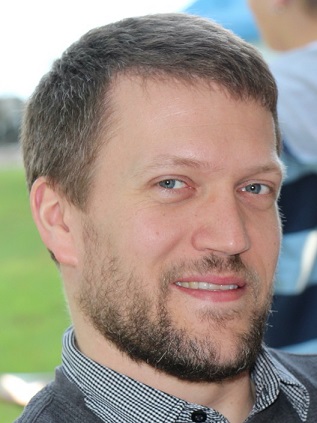}}]{Mathieu Salzmann}
is a Senior Researcher at EPFL. Previously, he was a Senior Researcher and Research Leader in NICTA's computer vision research group, a Research Assistant Professor at TTI-Chicago, and a postdoctoral fellow at ICSI and EECS at UC Berkeley. He obtained his PhD in Jan. 2009 from EPFL. His research interests lie at the intersection of machine learning and geometry for computer vision.
\end{IEEEbiography}

\vspace{-20pt}
\begin{IEEEbiography}[{\includegraphics[width=1in,height=1.95in,clip,keepaspectratio]{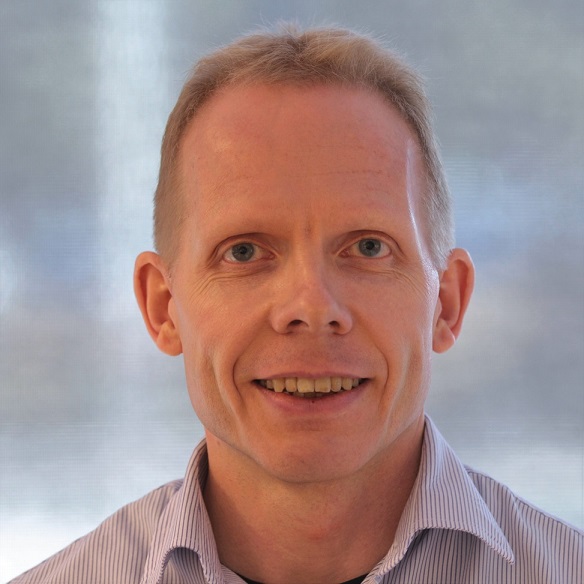}}]{Lars Petersson} is a Principal Research Scientist within Data61, CSIRO, Australia, specialising in resource constrained computer vision. Previously, he was a Principal Researcher in NICTA where he was leading projects such as Smart Cars, AutoMap, and Distributed Large Scale Vision. Earlier, he was a researcher at the Australian National University.  He received his PhD in 2002 from KTH, Stockholm, Sweden, where he also received his Master's degree in Engineering Physics.
\end{IEEEbiography}

\vspace{-20pt}
\begin{IEEEbiography}[{\includegraphics[width=1in,height=1.25in,clip,keepaspectratio]{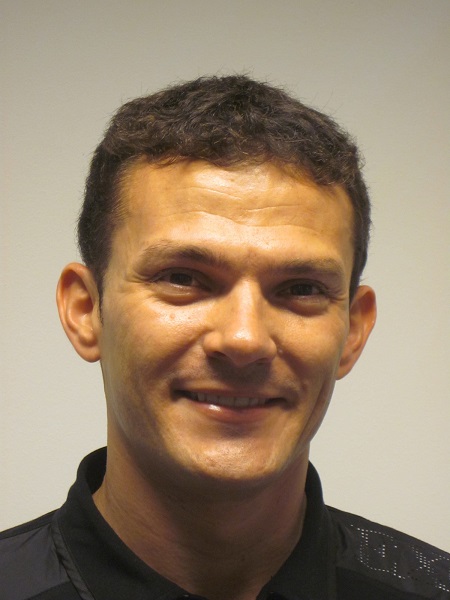}}]{Jose M. Alvarez} is a senior researcher at Data61, CSIRO, Australia working on deep learning and computer vision. He obtained his Ph.D. in 2010 from Autonomous University of Barcelona under the supervision of Prof. Antonio Lopez and Prof. Theo Gevers. Previous to CSIRO, he worked as a postdoctoral researcher at the Courant Institute of Mathematical Science at New York University under the supervision of Prof. Yann LeCun. In 2012, he joined NICTA as a computer vision researcher.
\end{IEEEbiography}

\vspace{-20pt}
\begin{IEEEbiography}[{\includegraphics[width=1in,height=1.25in,clip,keepaspectratio]{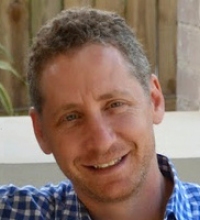}}]{Stephen Gould}
is an Associate Professor at ANU and Principal Research Scientist at Amazon. He is a former ARC Postdoctoral Fellow and Microsoft Faculty Fellow, and is a Contributed Researcher to the Data61 Machine Learning group and a Chief Investigator with the ARCV. Stephen received his BSc degree in mathematics and computer science and BE degree in electrical engineering from the University of Sydney in 1994 and 1996, respectively. He received his MS degree in electrical engineering from Stanford University in 1998. In 2005 he returned to PhD studies and earned his PhD degree in Electrical Engineering from Stanford University in 2010. His main research focus is on automatic semantic and geometric understanding of images and videos.
\end{IEEEbiography}




\end{document}